\documentclass[letterpaper]{article} 
\pdfoutput=1
\usepackage{aaai2026}  

\usepackage{times}  
\usepackage{helvet}  
\usepackage{courier}  
\usepackage[hyphens]{url}  
\usepackage{graphicx} 

\usepackage{natbib}  
\usepackage{caption} 
\usepackage{algorithm}
\usepackage{algorithmic}

\usepackage{newfloat}
\usepackage{listings}
\usepackage{latexsym}
\usepackage{times}
\usepackage{bm}
\usepackage{tabularx}
\usepackage{tabularray}
\usepackage{dsfont}

\usepackage{microtype}

\usepackage{inconsolata}

\usepackage{latexsym}
\usepackage{times}
\usepackage{bm}
\usepackage{tabularx}
\usepackage{tabularray}
\usepackage{dsfont}

\usepackage{microtype}

\usepackage{inconsolata}

\usepackage{microtype}

\usepackage{inconsolata}
\usepackage{algorithm}
\usepackage{algorithmic}
\usepackage{CJK}
\usepackage{colortbl}
\usepackage{inconsolata}
\usepackage{multirow}  
\usepackage{amssymb}   
\usepackage{amsmath}
\usepackage{fontawesome5}
\usepackage{subfigure} 
\usepackage{makecell} 
\usepackage{booktabs} 
\usepackage{enumitem} 
\usepackage{multirow} 
\usepackage[table]{xcolor}

\usepackage{booktabs} 
\usepackage{listings}
\usepackage{pifont}
\usepackage{tcolorbox}
\usepackage{soul}
\usepackage{enumitem}
\tcbuselibrary{skins, breakable, theorems}
\usepackage{lscape}
\usepackage{url}

\definecolor{mypink}{rgb}{.99,.91,.95}
\definecolor{myyellow}{rgb}{.99,.94,.82}
\definecolor{mygray}{gray}{.92}
\definecolor{tabcolor1}{RGB}{247,225, 237} 
\definecolor{darkgreen}{RGB}{50,100,0}
\definecolor{darkred}{RGB}{200, 0, 0}
\definecolor{lightred}{RGB}{250, 200, 200}
\definecolor{lightblue}{RGB}{210, 220, 250}
\newcommand{\cmark}{\textcolor{darkgreen}{\ding{51}}} %
\newcommand{\xmark}{\textcolor{darkred}{\ding{55}}} %
\definecolor{tabcolor2}{RGB}{255, 250, 132} 
\definecolor{tabcolor3}{RGB}{204, 232, 207} 
\definecolor{tabcolor4}{RGB}{245, 222, 179} 
\definecolor{tabcolor5}{RGB}{210, 220, 250} 
\definecolor{tabcolor6}{RGB}{222, 222, 222} 

\DeclareCaptionStyle{ruled}{labelfont=normalfont,labelsep=colon,strut=off} 
\floatstyle{ruled}
\newfloat{listing}{tb}{lst}{}
\floatname{listing}{Listing}
\title{Teaching Large Language Models to Maintain Contextual Faithfulness via Synthetic Tasks and Reinforcement Learning}
\author{
\textbf{Shuzheng Si\thanks{\ Equal Contribution.}$^{\spadesuit\diamondsuit}$, Haozhe Zhao\footnotemark[1]$^{\clubsuit}$, Cheng Gao\footnotemark[1]$^{\spadesuit}$, Yuzhuo Bai$^{\spadesuit}$, Zhitong Wang$^\spadesuit$} \\ \textbf{Bofei Gao$^{\heartsuit}$, Kangyang Luo$^{\spadesuit}$, Wenhao Li$^{\spadesuit}$, Yufei Huang$^{\spadesuit}$, Gang Chen$^{\diamondsuit}$} \\ 
\textbf{Fanchao Qi\thanks{\ Corresponding Authors.}$^{\spadesuit\diamondsuit}$, Minjia Zhang$^{\clubsuit}$, Baobao Chang$^{\heartsuit}$,} \textbf{Maosong Sun\footnotemark[2]$^{\spadesuit}$}
}
\affiliations{
$^{\spadesuit}$ Tsinghua University \quad $^{\heartsuit}$ Peking University
\quad $^\diamondsuit$ DeepLang AI \\
\quad $^{\clubsuit}$ University of Illinois Urbana-Champaign
}

\usepackage{bibentry}
\usepackage{arydshln}

\begin{document}

\maketitle

\begin{abstract}
Teaching large language models (LLMs) to be faithful in the provided context is crucial for building reliable information-seeking systems.
Therefore, we propose a systematic framework, \textsc{Canoe}, to reduce faithfulness hallucinations of LLMs across different downstream tasks without human annotations.
Specifically, we first synthesize short-form question-answering (QA) data with four diverse tasks to construct high-quality and easily verifiable training data without human annotation.
Also, we propose Dual-GRPO, a rule-based reinforcement learning method that includes three tailored rule-based rewards derived from synthesized short-form QA data, while simultaneously optimizing both short-form and long-form response generation.
Notably, Dual-GRPO eliminates the need to manually label preference data to train reward models and avoids over-optimizing short-form generation when relying only on the synthesized short-form QA data.
Experimental results show that \textsc{Canoe} greatly improves the faithfulness of LLMs across 11 different tasks, even outperforming the most advanced LLMs, e.g., GPT-4o and OpenAI o1. \footnote{~The data, code, and models will be available at \url{https://github.com/S1s-Z/CANOE}.
Email: ssz24@mails.tsinghua.edu.cn.}

\end{abstract}

\section{Introduction}
\label{section:introduction}
Recent progress in large language models (LLMs) has revolutionized text generation with their remarkable capabilities \citep{ OpenAI2023GPT4TR}.
In practice, LLMs are widely used to generate fluent and coherent text responses based on the provided contextual information, e.g., document question answering (QA) \citep{wang-etal-2024-leave} and text summarization \citep{zhang2024systematicsurveytextsummarization}.
However, LLMs often generate responses that are not faithful or grounded in the input context, i.e., faithfulness hallucinations \citep{hit-survey, si-etal-2025-aligning}, which can undermine their trustworthiness.
Maintaining faithfulness to the context is especially important in fields where accurate information transfer is essential \citep{duong2025scope}.
For instance, in legal summarization \citep{dong-etal-2025-termdiffusum}, the text output must reflect the content of legal documents without introducing any distortions.

\begin{figure}
    \centering
    \includegraphics[width=8cm]{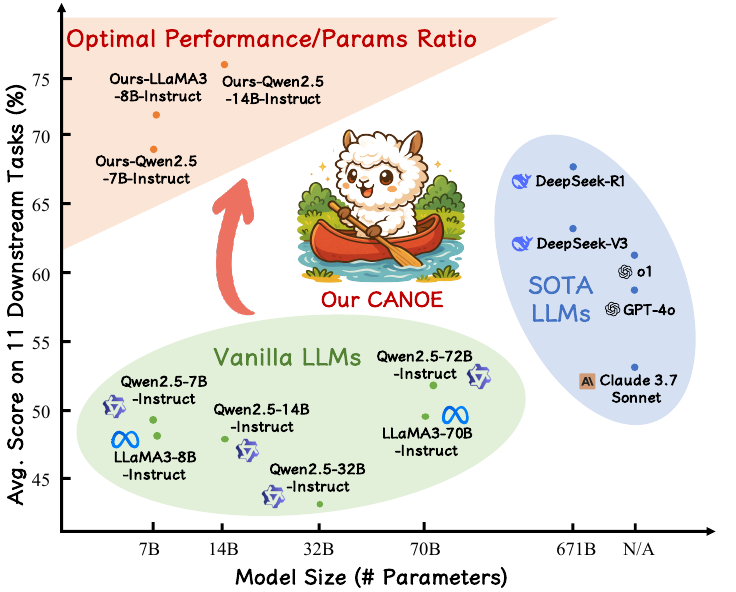}
    \caption{Average score on 11 downstream tasks vs model size. With only 7B parameters, \textsc{Canoe} already exceeds state-of-the-art LLMs like GPT-4o and  OpenAI o1.}
    \label{fig_intro}
\end{figure}

However, improving the faithfulness of LLMs faces three key challenges.
Specifically,
(1) \textbf{Faithfulness is difficult to improve by simply scaling model parameters}:
Previous works \citep{xie2024adaptive, li2025generatediscriminateevolveenhancing} find that LLMs may overly rely on internal knowledge learned from extensive pre-training data while disregarding provided contexts, i.e., the knowledge conflicts \citep{xu-etal-2024-knowledge-conflicts}.
When the model parameters increase and internal knowledge grows, this may lead to greater knowledge conflicts and further lower the faithfulness of LLMs \citep{ming2025faitheval}. 
Thus, it is necessary to explore a tailored post-training method to improve the faithfulness instead of simply scaling the model parameters.
(2) \textbf{Faithfulness is challenging to consistently boost across different downstream tasks}:
Recently, several methods \citep{li-etal-2024-improving-faithfulness, duong2025scope} have been proposed to improve the faithfulness of LLMs for different tasks.
For example, 
\citet{bi2024contextdpoaligninglanguagemodels} aligns LLMs through DPO \citep{dpo} with constructed faithful and unfaithful short-form completions, improving the performance of LLMs on short-form QA tasks.
However, these recent methods are designed for specific tasks, so they fail to consistently improve the faithfulness of LLMs across various tasks, like text summarization and multiple-choice questions, because these tasks can vary greatly.
\begin{figure*}[!h]
    \centering
    \vspace{-7.5mm}
    \includegraphics[width=0.97\linewidth]{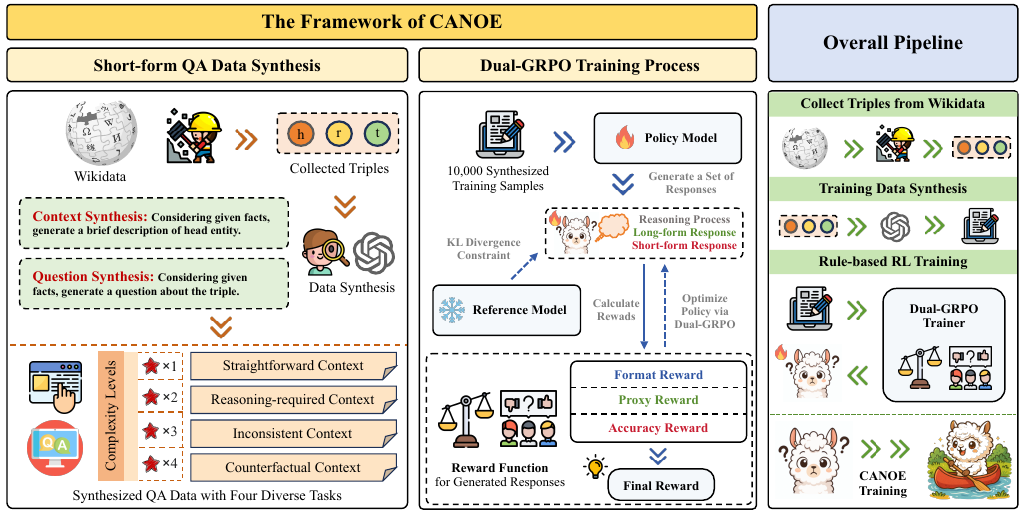}
    \caption{
     An overview of \textsc{Canoe} framework. \textsc{Canoe} first synthesizes easily verifiable short-form QA data and then proposes the Dual-GRPO with designed rule-based rewards to improve the faithfulness of LLMs. 
    }
    \label{figure:model}
\end{figure*}
(3) \textbf{Data used to enhance faithfulness is hard to scale}:
This issue is especially problematic with data used to improve the faithfulness in long-form generation tasks.
Unlike tasks with clear answers, e.g., short-form fact-seeking QA tasks \citep{wei2024measuringshortformfactualitylarge}, there is no standard way to ensure data quality in long-form generation tasks \citep{duong2025scope}.
Thus, data is typically annotated by humans \citep{zhu-etal-2023-annotating}, which is costly and not scalable.

To tackle these challenges, we propose a systematic post-training method called \textsc{Canoe}.
The main idea behind \textsc{Canoe} is to synthesize easily verifiable short-form QA data and then leverage reinforcement learning (RL) with tailored rule-based rewards to improve the faithfulness of LLMs in both short-form and long-form generation tasks.
\textsc{Canoe} firstly introduces Dual-GRPO, a variant of GRPO \citep{shao2024deepseekmathpushinglimitsmathematical} that includes three carefully tailored rule-based RL rewards derived from synthesized short-form QA data, while optimizing both short-form and long-form response generation.
For the provided contextual information and question, Dual-GRPO first prompts LLMs to produce a reasoning process, followed by a long-form answer composed of detailed and complete sentences, and finally a concise short-form and easily verifiable answer in just a few words.
For example, given the context, if the question is \textit{``What is the country of origin of Super Mario?''}, the long answer could be \textit{``Super Mario originated from Japan.''}, while the short answer could simply be \textit{``Japan''}.
In this way, we can assign different rewards to long-form and short-form responses, optimizing both simultaneously.
Note that we assign accuracy rewards on generated short-form responses since the short-form QA task enables reliable rule-based verification of faithfulness.
To overcome the problem of the faithfulness of the generated long-form responses being difficult to evaluate via rule-based verification \citep{ openai2025deep}, we propose proxy rewards to evaluate it implicitly.
Specifically, we construct the new input by replacing the given context with the generated long-form answer, then feed it to the LLMs to evaluate whether a long-form answer can drive the LLMs toward the correct short-form answer.
If the generated long-form response enables LLMs to generate the correct final answer, this indicates that it remains context-faithful and contains easy-to-understand sentences that answer the question correctly.
We also introduce format rewards to ensure more structured outputs and contribute to more stable training.
To obtain the data used for training without human annotation, we collect head-relation-tail triples from the knowledge base, apply the advanced GPT-4o \citep{OpenAI2023GPT4TR} to synthesize the question and contextual information, and use the tail entity from the triple as the answer to ensure the correctness.
Moreover, we introduce four diverse QA tasks to ensure the complexity and diversity of the training data.
Combined with the rule-based Dual-GRPO and data synthesis, \textsc{Canoe} can teach LLMs to remain context-faithful in both short-form and long-form generation tasks without relying on human annotations.

We evaluate the effectiveness of \textsc{Canoe} across 11 different downstream tasks, covering short-form and long-form generation tasks.
Results show that \textsc{Canoe} significantly reduces faithfulness hallucinations.
Specifically, \textsc{Canoe} significantly improves the overall score, e.g., 22.6\% for \textit{Llama3-Instruct-8B}.
Meanwhile, \textsc{Canoe} surpasses the most advanced LLMs (e.g., GPT-4o) in the overall score.
As shown in Figure \ref{fig_intro}, these results are unprecedented for open-source models that do not rely on additional human annotations.

\section{Methodology}
\label{section:method}

In this section, we will detail our proposed framework \textsc{Canoe}, which aims to teach LLMs to remain faithful across different tasks without human annotation.
Specifically, we first synthesize easily verifiable short-form QA data and then propose the Dual-GRPO with designed rule-based rewards to improve the faithfulness of LLMs in both short-form and long-form response generation. 
We start with the introduction of the short-form data synthesis process, then a brief overview of RL protocol, and the tailored rule-based rewards used in the proposed Dual-GRPO training.
An overview of the \textsc{Canoe} framework is presented in Figure \ref{figure:model}.

\subsection{Training Data Construction}
\label{sec:data}
Constructing high-quality and easily verifiable data is crucial for rule-based RL training \citep{shao2024deepseekmathpushinglimitsmathematical}.
Inspired by knowledge base question generation \citep{cui-etal-2019-kb, guo-etal-2024-diversifying}, we attempt to collect triples from the knowledge base and use the advanced LLMs to synthesize the context and question.
Concretely, we first collect about 30,000 head-relation-tail triples from Wikidata \citep{wikidata}.
Each collected triple $(h,r,t)$ includes a head entity $h$, a tail entity $t$, and the relation $r$ between two entities.
Then we craft prompt templates and query the most advanced GPT-4o to synthesize the contextual information $c$ and question $q$ based on the triple $(h,r,t)$.
We directly use the tail entity $t$ as the final answer $a$ to ensure the correctness and easy validation of the synthesized data.
Each synthetic short-form QA sample $(c, q, a)$ consists of a contextual passage $c$, a question $q$, and a ground truth answer $a$.
In this way, we can obtain short-form QA data that can be easily verified; thus, we can utilize a rule-based RL method to optimize our LLMs to be more faithful.
Meanwhile, to ensure the complexity and diversity of training data, we design four diverse QA tasks, including straightforward context, reasoning-required context, inconsistent context, and counterfactual context.
The model is expected to answer the question by leveraging the information in the provided context.

\noindent
\textbf{Straightforward Context.}
A straightforward context means that the context clearly contains statements of the final answer.
It requires models to accurately locate and utilize information from the context in order to answer questions.
Specifically, we keep the original collected triple as input to query GPT-4o to synthesize the data $(c, q, a)$.

\noindent
\textbf{Reasoning-required Context.}
This type of context contains multiple related entities and relations, and requires models to correctly answer multi-hop reasoning questions.
Firstly, we construct a subgraph based on the sampled triples and extract 2, 3, 4-hop paths $[(h^1, r^1, t^1),...,(h^n, r^n, t^n)]_{n\leq4}$.
Then, we use the $n$-th tail entity $t^n$ as the ground truth answer and employ the constructed paths to query GPT-4o to obtain the multi-hop context and question.

\noindent
\textbf{Inconsistent Context.}
This involves multiple randomly ordered contexts generated from different triples.
This simulates noisy and inconsistent scenarios, where models need to detect inconsistencies and focus on useful and relevant contexts to answer the questions.
We construct such a sample by combining the contexts from up to three QA samples.

\noindent
\textbf{Counterfactual Context.}
A counterfactual context contains statements that contradict common sense within the collected triples.
Firstly, we replace the tail entity $t$ of the original collected triple with a similar but counterfactual entity  $t^{cf}$.
Then, we query GPT-4o to generate questions and counterfactual contexts to construct counterfactual samples.
Unlike the aforementioned tasks, this task further highlights the importance of faithfulness for LLMs to answer the questions correctly, as it prevents models from depending on their learned factual knowledge to find the right answers.

By introducing four different tasks, we construct 10,000 QA pairs used for training without human annotation.
These short-form QA data can be easily verified and include tasks varying in complexity, which can make rule-based RL training more efficient in improving the faithfulness.
More details are shown in the Appendix~\ref{appendix:traning_data}, e.g., used prompts, data mixing recipes, and data statistics.

\subsection{Reinforcement Learning Protocol}
For RL training of LLMs, methods based on policy optimization, such as PPO \citep{ppo} and GRPO \citep{shao2024deepseekmathpushinglimitsmathematical}, have been explored. 
Given the effectiveness of GRPO in training models and its advantages over PPO, e.g., eliminating the need for human-annotated preference data to train a reward model, we utilize GRPO to optimize and improve the faithfulness of the policy model $\pi_{\theta}$.

For each input, consisting of provided contextual information $c$, a natural language question $q$, the model generates a group of $G$ candidate answers, $\{o_1, o_2, \dots, o_G\}$. 
Each candidate is evaluated using a designed composite rule-based reward function to capture the end goal of faithfulness.
GRPO leverages the relative performance of candidates within the group to compute an advantage $A_i$ for each output, guiding policy updates according to the following objective:
\begin{equation}		
\label{grpo_1}
		\resizebox{0.89\hsize}{!}{$\begin{aligned}
            \mathcal{J}_\text{GRPO}(\theta) &= \mathbb{E}_{c, q, \{o_i\} \sim \pi_{\theta_{old}}} \left[ \frac{1}{G} \sum_{i=1}^G \mathcal{L}_i - \beta \mathbb{D}_{KL}(\pi_{\theta} || \pi_{ref}) \right],
		\end{aligned}$}
\end{equation}
\begin{equation}		
\label{grpo_2}
		\resizebox{0.75\hsize}{!}{$\begin{aligned}
            \mathcal{L}_i &= \min \left( w_i A_i, \text{clip}(w_i, 1 - \epsilon, 1 + \epsilon) A_i \right),
		\end{aligned}$}
\end{equation}

\noindent
where $w_i = \frac{\pi_\theta(o_i |q)}{\pi_{\theta_{old}}(o_i |q)}$, $\pi_{\theta_{\text{old}}}$ is the policy before the update, $\pi_{\text{ref}}$ is the reference policy (i.e., the initial model), $\epsilon$ and $\beta$ are hyperparameters controlling the update step and divergence regularization and $A_i$ is computed using the normalized reward within the group. 
We use our synthesized QA data as training data, which is easily verifiable, so that we can apply GRPO and train LLMs using rule-based rewards.
Also, employing the rule-based GRPO removes the need for humans to label preference data for training the reward model.

\subsection{Reward Design}
Having a well-designed reward is key to the effectiveness of RL training \citep{ kimiteam2025kimik15scalingreinforcement}. 
To use easily verifiable short-form QA data to improve the faithfulness of LLMs, the most intuitive reward would be the accuracy reward, which can check if the generated responses match the ground truth answers. 
However, in our early experiments, we found that relying solely on short-form QA data and accuracy rewards fails to enhance the faithfulness of long-form response generation, as the models may over-optimize short-form generation and learn a false pattern.
The tuned models tend to copy text spans from the context as answers and lose their ability to generate long-form responses.
Meanwhile, directly evaluating the faithfulness of long, free-form responses via the rule-based verification continues to pose an unresolved challenge.


Therefore, we propose \textbf{Dual-GRPO}, which includes a set of well-designed rewards that provide more harmonized guidance for optimizing LLMs to generate faithful responses. 
Unlike the original GRPO that over-optimizes short-form generation, we first prompt LLMs to generate both long-form and short-form responses, then assign different rewards to the two generated responses to improve the faithfulness.

\noindent
\textbf{System Prompt and Rollouts.}
For the provided context and question, we introduce a system prompt that requires LLMs to produce a reasoning process, then a long-form answer composed of detailed and complete sentences, and finally a concise short-form answer in just a few words.
For example, given the context, if the question is \textit{``What is the country of origin of Super Mario?''}, the long answer could be \textit{``Super Mario originated from Japan.''}, while the short answer could simply be \textit{``Japan''}.
In this way, we can assign different reward scores to long-form and short-form answers while optimizing them both at once.
This system prompt also triggers zero-shot chain-of-thought reasoning in the policy model, which progressively improves as training advances to optimize for the reward.
The system prompt used for Dual-GRPO rollouts is shown in the Appendix~\ref{appendix:grpo}.

\noindent
\textbf{Accuracy Reward for Short-form Response Generation.}
This reward directly assesses whether the generated short-form responses match the ground truth answers.
We use the exact matching (EM) to measure accuracy, giving a score of 1 for a match and 0 for a mismatch. 
Thus, we can ensure that the generated short-form response correctly answers the question based on the context, making LLMs more faithful in short-form response generation.

\noindent
\textbf{Proxy Reward for Long-form Response Generation.}
Evaluating the faithfulness of long-form responses via the rule-based verification remains challenging.
This is because these long-form answers are often free-form, making rule-based verification ineffective \citep{openai2025deep}.
Therefore, instead of directly evaluating the faithfulness of the long-form response, we propose a proxy reward to evaluate it implicitly, as the faithfulness of a long-form answer can be measured by its ability to drive the LLMs toward a correct short-form answer.
Specifically, for each generated long-form answer $y_{lf}$, we replace the given context $c$ with it as new input and feed it to the LLM to check whether the LLM can produce the correct short-form answer based on $y_{lf}$.
If the generated long-form response can enable the LLM to generate the correct answer, it indicates that the long-form response stays faithful to the context, contains complete and easy-to-understand sentences, and correctly addresses the question.
Thus, we assign a reward score of 1 for the positive long-form response that helps the LLM to produce the correct final answer, and a reward score of 0 for those that lead to an incorrect answer.

\noindent
\textbf{Format Reward.}
We also include a format reward that encourages adherence to a predefined output structure (e.g., using \textless think\textgreater, \textless long\_answer\textgreater, and \textless short\_answer\textgreater tags).
Outputs that conform to this pattern receive a reward boost, thereby enhancing consistency. 
We use string matching to evaluate whether the generated responses adhere to the format, giving a score of 1 for a match and 0 for a mismatch.

Finally, we use the sum of these three rewards as the final composite reward.
It enhances the efficacy of the RL training framework, guiding the model toward generating more faithful responses in both short-form and long-form tasks.
More details are shown in the Appendix~\ref{appendix:grpo}.

\section{Experiments}
In this section, we conduct experiments and provide analyses to justify the effectiveness of \textsc{Canoe}.

\begin{table*}
\centering  
\vspace{-3.5mm}
\resizebox{1\hsize}{!}{
\begin{tabular}{lcccccccccccccc}
\toprule
\multicolumn{1}{l}{\multirow{3}{*}{\textbf{Model}}} & \multicolumn{9}{c}{\textbf{Short-form Generation Tasks}} & \multicolumn{3}{c}{\textbf{Long-form Generation Tasks}} &  \multicolumn{2}{c}{\multirow{2}{*}{\textbf{Avg. Score}}}\\
\cmidrule(lr){2-10} \cmidrule(lr){11-13} & \multicolumn{2}{c}{\textbf{ConFiQA}} & \multicolumn{2}{c}{\textbf{FiQA}} & \multicolumn{2}{c}{\textbf{CNQ}} & \multicolumn{1}{c}{\textbf{FaithEval}}  & \multicolumn{2}{c}{\textbf{FollowRAG}} & \multicolumn{1}{c}{\textbf{XSum}} & \multicolumn{1}{c}{\textbf{WikiLarge}} & \multicolumn{1}{c}{\textbf{CLAPNQ}} \\
\quad &  \textbf{EM} & \textbf{Acc} & \textbf{EM}   & \textbf{Acc}  & \textbf{EM} & \textbf{Acc}  & \textbf{Acc}  & \textbf{EM} &\textbf{Acc} & \textbf{FS} & \textbf{FS} & \textbf{FS} & \textbf{Avg EM} & \textbf{Avg Acc} \\
\midrule
\rowcolor{mygray} \multicolumn{15}{c}{\cellcolor{myyellow} \textbf{The state-of-the-art LLMs}} \\
GPT-4o & 31.5 & 42.7 & 66.8 & 79.6 & 43.4 & 55.9 & 47.5 & 42.2 & 57.8 & 80.7 & 88.1 & 70.3 & 58.8 & 65.3 \\
GPT-4o mini & 49.5 & 63.7 & 67.1 & 78.8 & 47.8 & 54.3 & 50.9 & 38.5 & 51.3 & 75.4 & 91.0 & 66.0 & 60.8 & 66.4 \\
DeepSeek V3 & 49.5 & 58.6 & 67.0 & 76.5 & 54.6 & 67.3 & 51.0 & 37.7 & 55.2 & 82.8 & 85.6 & 71.0 & 62.4 & 68.5 \\
Claude 3.7 Sonnet & 26.0 & 36.0 & 56.4 & 72.2 & 41.4 & 65.0 & 45.6 & 36.3 & 53.7 & 78.3 & 81.7 & 68.3 & 54.3 & 62.6 \\
\hdashline[2pt/3pt]
OpenAI o1 & 49.0 & 57.9 & 78.0 & 89.7 & 29.5 & 39.1 & 52.0 & 40.5 & 57.0 & 81.0 & 88.1 & 68.0 & 60.8 & 66.6 \\
DeepSeek R1 & 68.4 & 74.3 & 68.4 & 80.7 & 60.3 & 70.2 & 60.1 & 42.9 & 56.6 & 80.3 & 83.0 & 73.5 & 67.1 & 72.3 \\
Claude 3.7 Sonnet-Thinking & 27.1 & 38.7 & 59.5 & 76.7 & 42.1 & 67.0 & 57.0 & 38.8 & 55.3 & 79.0 & 81.4 & 72.2 & 57.1 & 65.9 \\
\midrule
\rowcolor{mygray} \multicolumn{15}{c}{\cellcolor{myyellow} \textbf{LLaMA-3-Instruct Series}} \\
LLaMA-3-Instruct-8B & 49.2 & 58.2 & 11.4 & 59.3 & 37.8 & 45.2 & 52.0 & 31.1 & 44.8 & 64.2 & 77.1 & 58.5 & 47.7 & 57.4 \\
LLaMA-3-Instruct-70B & 38.1 & 54.5 & 9.1 & 66.8 & 54.2 & 65.0 & 50.9 & 38.7 & 45.7 & 72.0 & 77.4 & 47.2 & 48.5 & 59.9 \\
\hdashline[2pt/3pt]
SFT-8B & 65.1 & 70.3 & 35.9 & 59.9 & 52.6 & 65.7 & 43.0 & 19.2 & 21.0 & 62.2 & 74.2 & 55.3 & 50.9 & 56.4 \\
Context-DPO-8B & 66.3 & 72.9 & 40.9 & 59.5 & 54.6 & 62.3 & 37.5 & 29.9 & 43.8 & 65.2 & 78.2 & 59.1 & 54.0 & 59.8 \\
SCOPE$_{sum}$-8B & 35.7 & 64.6 & 7.1 & 68.7 & 33.8 & 60.6 & 55.7 & 30.1 & 46.2 & 70.3 & 80.3 & 59.8 & 46.6 & 63.3 \\
\rowcolor{blue!5} \textbf{\textsc{Canoe}-LLaMA-8B} & \textbf{73.5} & \textbf{80.9} & \textbf{82.7} & \textbf{84.9} & \textbf{66.7} & \textbf{73.4} & \textbf{74.6} & \textbf{40.9} & \textbf{51.7} & \textbf{74.4} & \textbf{84.4} & \textbf{64.9} & \textbf{70.3} & \textbf{73.6} \\
\hdashline[2pt/3pt]
\rowcolor{blue!5}  $\Delta$ Compared to Vanilla. & \textcolor[rgb]{0.7,0,0}{+24.3} & \textcolor[rgb]{0.7,0,0}{+22.6} & \textcolor[rgb]{0.7,0,0}{+71.3} & \textcolor[rgb]{0.7,0,0}{+25.6} & \textcolor[rgb]{0.7,0,0}{+28.9} & \textcolor[rgb]{0.7,0,0}{+28.2} & \textcolor[rgb]{0.7,0,0}{+22.6} & \textcolor[rgb]{0.7,0,0}{+9.8} & \textcolor[rgb]{0.7,0,0}{+6.9} & \textcolor[rgb]{0.7,0,0}{+10.2} & \textcolor[rgb]{0.7,0,0}{+7.3} & \textcolor[rgb]{0.7,0,0}{+6.4} & \textcolor[rgb]{0.7,0,0}{+22.6} & \textcolor[rgb]{0.7,0,0}{+16.2} \\
\midrule
\rowcolor{mygray} \multicolumn{15}{c}{\cellcolor{myyellow} \textbf{Qwen-2.5-Instruct Series}} \\
Qwen-2.5-Instruct-7B & 52.5 & 61.0 & 13.2 & 68.4 & 55.3 & 68.2 & 56.1 & 32.6 & 45.3 & 63.4 & 57.8 & 61.2 & 49.0 & 60.2 \\
Qwen-2.5-Instruct-14B & 34.1 & 47.3 & 0.8 & 61.4 & 43.1 & 64.3 & 51.6 & 34.8 & 51.2 & 68.2 & 82.3 & 63.4 & 47.3 & 61.2 \\
Qwen-2.5-Instruct-32B & 44.5 & 66.4 & 39.2 & 81.1 & 37.7 & 66.4 & 47.0 & 33.9 & 53.1 & 20.2 & 57.7 & 31.7 & 39.0 & 52.9 \\
Qwen-2.5-Instruct-72B & 43.7 & 52.3 & 4.8 & 67.3 & 51.8 & 62.2 & 45.2 & 38.5 & 55.7 & 71.2 & 90.4 & 64.8 & 51.3 & 63.6 \\
\hdashline[2pt/3pt]
SFT-7B & 62.8 & 69.8 & 48.8 & 76.6 & 60.1 & 65.3 & 50.3 & 29.0 & 41.7 & 55.2 & 51.3 & 57.2 & 51.8 & 58.4 \\
Context-DPO-7B & 64.5 & 70.6 & 57.1 & 78.2 & 62.3 & 70.1 & 45.7 & 31.0 & 43.7 & 60.2 & 53.4 & 62.8 & 54.6 & 60.6 \\
SCOPE$_{sum}$-7B & 39.3 & 47.9 & 12.9 & 60.9 & 50.2 & 55.3 & 52.3 & 30.6 & 46.0 & 68.3 & 72.0 & 63.2 & 48.6 & 58.2 \\
\rowcolor{blue!5}  \textbf{\textsc{Canoe}-Qwen-7B}& \textbf{67.6} & \textbf{75.2} & \textbf{78.1} & \textbf{83.5} & \textbf{67.2} & \textbf{76.4} & \textbf{70.5} & \textbf{37.0} & \textbf{50.2} & \textbf{72.4} & \textbf{86.1} & \textbf{65.2} & \textbf{68.0} & \textbf{72.4} \\
\hdashline[2pt/3pt]
\rowcolor{blue!5}  $\Delta$ Compared to Vanilla. & \textcolor[rgb]{0.7,0,0}{+15.1} & \textcolor[rgb]{0.7,0,0}{+14.2} & \textcolor[rgb]{0.7,0,0}{+64.9} & \textcolor[rgb]{0.7,0,0}{+15.0} & \textcolor[rgb]{0.7,0,0}{+11.9} & \textcolor[rgb]{0.7,0,0}{+8.2} & \textcolor[rgb]{0.7,0,0}{+14.4} & \textcolor[rgb]{0.7,0,0}{+4.4} & \textcolor[rgb]{0.7,0,0}{+4.9} & \textcolor[rgb]{0.7,0,0}{+9.0} & \textcolor[rgb]{0.7,0,0}{+28.3} & \textcolor[rgb]{0.7,0,0}{+4.0} & \textcolor[rgb]{0.7,0,0}{+19.0} & \textcolor[rgb]{0.7,0,0}{+12.3} \\
\hdashline[2pt/3pt]
\rowcolor{blue!5}  \textbf{\textsc{Canoe}-Qwen-14B} & \textbf{85.7} & \textbf{87.4} & \textbf{87.8} & \textbf{88.5} & \textbf{81.8} & \textbf{84.2} & \textbf{67.4} & \textbf{46.1} & \textbf{54.6} & \textbf{75.7} & \textbf{91.1} & \textbf{68.4} & \textbf{75.5} & \textbf{77.2} \\
\rowcolor{blue!5}  $\Delta$ Compared to Vanilla. & \textcolor[rgb]{0.7,0,0}{+51.6} & \textcolor[rgb]{0.7,0,0}{+40.1} & \textcolor[rgb]{0.7,0,0}{+87.0} & \textcolor[rgb]{0.7,0,0}{+27.1} & \textcolor[rgb]{0.7,0,0}{+38.7} & \textcolor[rgb]{0.7,0,0}{+19.9} & \textcolor[rgb]{0.7,0,0}{+15.8} & \textcolor[rgb]{0.7,0,0}{+11.3} & \textcolor[rgb]{0.7,0,0}{+3.4} & \textcolor[rgb]{0.7,0,0}{+7.5} & \textcolor[rgb]{0.7,0,0}{+8.8} & \textcolor[rgb]{0.7,0,0}{+5.0} & \textcolor[rgb]{0.7,0,0}{+28.2} & \textcolor[rgb]{0.7,0,0}{+16.0} \\
\bottomrule
\end{tabular}}
\caption{Experimental results (\%) on eleven datasets. 
The FollowRAG results represent the results averaged over these four open-domain QA datasets, as shown in the Appendix~\ref{appendix:eval}, including NaturalQA, TriviaQA, HotpotQA, and WebQSP.
\textbf{Bold} numbers indicate the best performance of models with the same model size.
Avg EM/Acc represents the average score between short-form task metrics (EM/Acc) and long-form task metric FaithScore (FS).
}
\label{tb:main}
\end{table*}

\subsection{Tasks and Datasets}

To evaluate our method \textsc{Canoe} comprehensively, we select a range of downstream datasets, including short-form and long-form generation tasks.

\noindent
\textbf{Short-form Generation Tasks.} 
For short-form generation tasks, we use two counterfactual QA datasets, including ConFiQA \citep{bi2024contextdpoaligninglanguagemodels} and CNQ \citep{longpre-etal-2021-entity}, a multiple-choice questions dataset FaithEval \citep{ming2025faitheval}, and a factual QA dataset FiQA \citep{bi2024contextdpoaligninglanguagemodels} that is the factual version of ConFiQA.
These datasets ensure the answers appear in the contexts to evaluate the faithfulness.
We also evaluate our method on four open-domain QA datasets within the FollowRAG benchmark \citep{dong2024generalinstructionfollowingalignmentretrievalaugmented} to evaluate the abilities of LLMs in real-world retrieval-augmented generation (RAG) scenarios, including NaturalQA \citep{47761}, TriviaQA \citep{joshi-etal-2017-triviaqa}, HotpotQA \citep{yang-etal-2018-hotpotqa}, and WebQSP \citep{Yih2016TheVO}.
In real-world RAG scenarios, the answer may not appear in the retrieved passages, and these passages tend to be noisy.
We evaluate models based on whether gold answers are included in the generated responses (i.e., Acc) following \citet{asai2024selfrag} and exact matching (EM) for QA tasks.
For multiple-choice questions, we follow \citet{ming2025faitheval} and use keyword matching to verify the accuracy.

\noindent
\textbf{Long-form Generation Tasks.} 
We include a text summarization task XSum \citep{narayan-etal-2018-dont}, a text simplification task WikiLarge \citep{zhang-lapata-2017-sentence}, and a long-form QA task CLAPNQ \citep{rosenthal-etal-2025-clapnq}.
To evaluate the faithfulness of generated long-form answers, called FaithScore (FS), we use MiniCheck \citep{tang-etal-2024-minicheck} to check whether the model response is grounded in the provided context.
MiniCheck is a state-of-the-art method to recognize if LLM output can be grounded in given contexts.
If the model response contains at least one statement that cannot be inferred from the context, we consider it as a negative response; otherwise, it is a positive response.
We also query GPT-4o to evaluate the quality of generated responses, namely QualityScore.

More details can be found in the Appendix~\ref{appendix:eval}.

\subsection{Baselines and Implementation Details}
\noindent
\textbf{Baselines.} 
We compare several baselines, including (1) \textbf{Vanilla LLMs}: including LLaMA-3-Instruct \citep{llama3} and Qwen-2.5-Instruct \citep{qwen2.5} of different sizes. 
We also conduct supervised fine-tuning on synthesized 10,000 short-form data as SFT baselines.
(2) \textbf{SOTA LLMs}: 
We further evaluate the most advanced LLMs, including GPT-4o, GPT-4o-mini, OpenAI o1 \citep{jaech2024openai}, Claude 3.7 Sonnet \citep{Claude3S}, Claude 3.7 Sonnet-Thinking, DeepSeek R1, and DeepSeek V3 \citep{deepseekai2025deepseekr1incentivizingreasoningcapability, deepseekai2025deepseekv3technicalreport}.
(3) \textbf{The Methods to Improve Faithfulness}:
Context-DPO \citep{bi2024contextdpoaligninglanguagemodels} aligns LLMs through DPO with constructed faithful and unfaithful short-form answers, thus improving the faithfulness in short-form generation.
SCOPE \citep{duong2025scope} proposes a pipeline to generate self-supervised task-specific data and applies preference training to enhance the faithfulness in a special task.
We train it on the sampled training set of the summarization task XSum as SCOPE$_{sum}$, regarding it as the method designed to improve the faithfulness of long-form response generation. 

\noindent
\textbf{Implementation Details.}
Our experiments are conducted on LLaMA-3-Instruct and Qwen-2.5-Instruct. 
Details are shown in Appendix~\ref{appendix:implementation}, e.g., hyperparameters.

\subsection{Main Results}

\noindent
\textbf{\textsc{Canoe} Improves the Faithfulness of LLMs in Both Short-form and Long-form Response Generation.}
As shown in Table \ref{tb:main}, \textsc{Canoe} shows significant improvements on 11 faithfulness-related benchmarks.
\textsc{Canoe} achieves substantial improvements in the overall score compared to original LLMs, e.g., \textbf{22.6\%} for \textit{Llama3-8B} and \textbf{19.0\%} for \textit{Qwen2.5-7B} in Avg EM score.
\textsc{Canoe} also surpasses the most advanced LLMs (e.g., GPT-4o) in the overall score (both Avg EM and Avg Acc scores).
This shows that \textsc{Canoe} effectively aligns LLMs to be context-faithful.
Also, for real-world RAG scenarios, our proposed \textsc{Canoe} can also improve the performance even though the answer may not appear in the retrieved passages, and these passages are often noisy.

\begin{table}[t]
\scriptsize	
\centering
\resizebox{0.93\linewidth}{!}{
\begin{tabular}{lcccc}
\toprule
\textbf{Model} & \textbf{XSum} & \textbf{WikiLarge} & \textbf{CLAPNQ} & \textbf{Avg} \\
\midrule
GPT-4o & 98.5 &	97.5 &	81.2 &	92.4  \\
\hdashline[2pt/3pt]
LLaMA-3-Instruct-8B & 70.9 &	82.9 &	39.2 	&64.3  \\
LLaMA-3- Instruct-70B  & 86.2 &	83.0 &	30.1 	&66.4  \\
\rowcolor{blue!5} \textbf{\textsc{Canoe}-LLaMA-8B} & \textbf{85.8} 	& \textbf{87.8} 	& \textbf{65.5} &	\textbf{79.7}  \\
\hdashline[2pt/3pt]
Qwen-2.5-Instruct-7B  & 79.4 	&79.0 	&64.6 	&74.3  \\
Qwen-2.5-Instruct-14B  & 90.5 	&83.1 	&63.6 	&79.1  \\
Qwen-2.5-Instruct-32B  & 90.3 	&83.9 &	58.6 	&77.6  \\
Qwen-2.5-Instruct-72B  & 95.7 	&94.1 	&75.4 	&88.4  \\
\hdashline[2pt/3pt]
\rowcolor{blue!5} \textbf{\textsc{Canoe}-Qwen-7B} & \textbf{91.5} 	& \textbf{87.3} 	& \textbf{68.2} 	& \textbf{82.3} \\
\rowcolor{blue!5} \textbf{\textsc{Canoe}-Qwen-14B} & \textbf{91.9} 	& \textbf{89.7} 	& \textbf{73.5} & 	\textbf{85.0} \\
\bottomrule
\end{tabular}}
\caption{QualityScore (\%) on long-form generation tasks.}
\label{tb:quality} 
\end{table}

\begin{table}[t]
\scriptsize	
\centering
\resizebox{0.93\linewidth}{!}{
\begin{tabular}{lcccccc}
\toprule
\multicolumn{1}{l}{\multirow{2}{*}{\textbf{Model}}} & \multicolumn{3}{c}{\textbf{Acc}} & \multicolumn{3}{c}{\textbf{EM}} \\
\cmidrule(lr){2-4} \cmidrule(lr){5-7} & QA & MR & MC   & QA & MR & MC \\
\midrule
GPT-4o & 52.2 &	45.6 &	30.3& 43.3 &	32.4 	& 18.7  \\
\hdashline[2pt/3pt]
LLaMA-3-Instruct-8B  &69.7 	&55.9 &	49.1& 60.0 	&47.9& 	39.6  \\
\rowcolor{blue!5} \textbf{\textsc{Canoe}-LLaMA-8B} & \textbf{82.7} &	\textbf{80.1} &	\textbf{79.8} & \textbf{76.4} &		\textbf{73.5} 	&	\textbf{70.5}   \\
\hdashline[2pt/3pt]
Qwen-2.5-Instruct-7B & 72.8 &	59.1 &	51.1 & 64.9 &		50.2 &		42.5  \\
Qwen-2.5-Instruct-14B & 62.4 &	44.9 	&34.7 &	 44.7 &		34.3 &		23.3  \\
Qwen-2.5-Instruct-32B  &74.1 &	65.9 	&59.3  &55.9 	&42.8 	&34.8 \\
Qwen-2.5-Instruct-72B & 63.3 &	50.3 &	43.3  &54.3 &	42.2 &	34.7 \\
\hdashline[2pt/3pt]
\rowcolor{blue!5} \textbf{\textsc{Canoe}-Qwen-7B} & \textbf{79.5} &	\textbf{76.1} &\textbf{70.1}	& \textbf{73.3} &		\textbf{67.9} &		\textbf{61.7}   \\
\rowcolor{blue!5} \textbf{\textsc{Canoe}-Qwen-14B} & \textbf{91.8} 	& \textbf{86.4} 	& \textbf{84.1} & \textbf{89.7} &		\textbf{85.2} &		\textbf{82.1}  \\
\bottomrule
\end{tabular}}
\caption{Results (\%) on three tasks in ConFiQA.}
\label{tb:3_tasks} 
\end{table}

\begin{figure*}[t]
    \centering
    \includegraphics[width=0.82\linewidth]{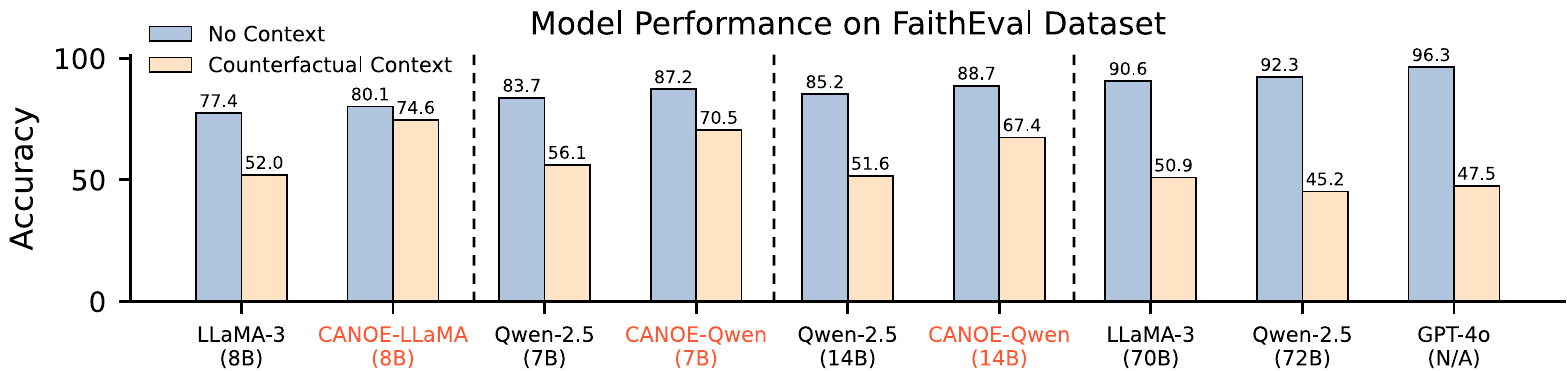}
    \caption{
    Model performance comparison on FaithEval in a closed-book QA setting and counterfactual context setting.
    Our models are colored in orange.
    We report the results from the chat version of LLaMA-3 and Qwen-2.5.
    }
    \label{figure:factuality}
\end{figure*}

\noindent
\textbf{\textsc{Canoe} Maintains the Factuality of LLMs.}
We further evaluate whether \textsc{Canoe} will reduce the factuality of LLMs.
Following \citet{ming2025faitheval}, we modify the original FaithEval and make it a closed-book QA setting, where no context is provided and LLMs need to give factual answers.
In this case, the models rely entirely on their parametric knowledge of common facts, and we find that our proposed \textsc{Canoe} maintains the factuality compared to the untuned LLM as shown in Figure \ref{figure:factuality}.
However, when a new context with counterfactual evidence that contradicts the model’s parametric knowledge is introduced, performance declines sharply.
For example, GPT-4o achieves 96.3\% accuracy on factual closed-book QA task but only 47.5\% on counterfactual QA task that evaluates the faithfulness of LLMs.
This highlights that, unlike factuality, the faithfulness of LLMs is difficult to improve by simply scaling model parameters, which further indicates the necessity of a post-training method to improve faithfulness.

\noindent
\textbf{\textsc{Canoe} Improves the Quality of Long-form Response Generation.}
As shown in Table \ref{tb:quality}, we can find that our proposed \textsc{Canoe} consistently improves the generation quality in the three long-form tasks.
This is because the proxy reward implicitly requires LLMs to generate easy-to-understand responses, which further optimizes the response quality.

\noindent
\textbf{\textsc{Canoe} Enhances LLMs’ Reasoning in Short-form Response Generation.}
ConFiQA consists of three different tasks: question answering (QA), multi-hop reasoning (MR), and multi-conflicts reasoning (MC). 
QA focuses on the single-hop task with context containing one corresponding answer, while MR and MC involve multi-hop reasoning tasks with context containing one and multiple related counterfactual contexts, respectively.
As shown in Table \ref{tb:3_tasks}, \textsc{Canoe} not only improves the faithfulness in the single-hop QA task but also enhances the reasoning ability in reasoning tasks.

\begin{figure}
    \centering
    \includegraphics[width=6.8cm]{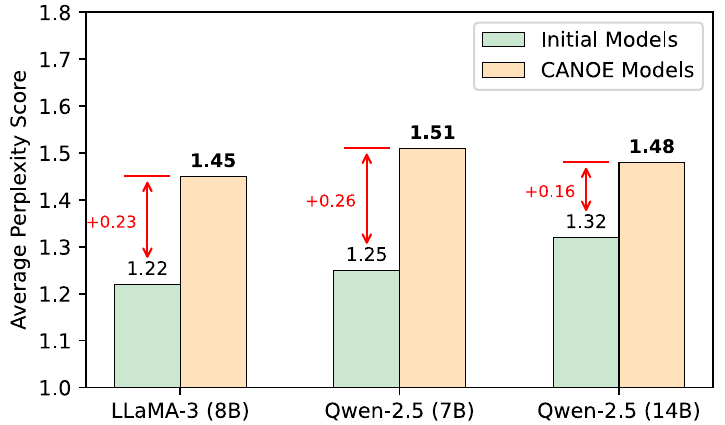}
    \caption{The average perplexity score of 110 negative samples for each model from eleven datasets.}
    \label{fig_overconfidence}
\end{figure}

\noindent
\textbf{\textsc{Canoe} Mitigates Overconfidence Bias.}
For each model, we select a total of 110 unfaithful samples with the highest perplexity from the 11 datasets, 10 samples per dataset.
Then we report the average perplexity score on these negative samples shown in Figure \ref{fig_overconfidence}.
We can find that \textsc{Canoe} produces the high perplexity scores, indicating low confidence scores, for these bad cases. 
This shows that \textsc{Canoe} mitigates overconfidence in these false statements.

\begin{table}[t]
\scriptsize	
\centering
\resizebox{\linewidth}{!}{
\begin{tabular}{lcccc}
\toprule
\multicolumn{1}{l}{\multirow{2}{*}{\textbf{Model}}} & \multicolumn{2}{c}{\textbf{Short-form Tasks}} & \multicolumn{2}{c}{\textbf{Long-form Tasks}} \\
\cmidrule(lr){2-3} \cmidrule(lr){4-5} & EM & Acc & FaithScore & QualityScore \\
\midrule
\rowcolor{blue!5} \textbf{\textsc{Canoe}-LLaMA-8B} &\textbf{67.7} & \textbf{73.1} & \textbf{74.6} & \textbf{79.7} \\
-w/o. Dual-GRPO \& Data Synthesis  & 36.3  & 51.9  & 66.6 & 64.3 \\
\hdashline[2pt/3pt]
-w/o. Dual-GRPO (i.e., original GRPO) & 60.5 & 66.6 & N/A & 23.5 \\
\hdashline[2pt/3pt]
-w/o. Reasoning-required Context. & 63.7 & 69.4 & 71.7 & 75.3 \\
-w/o. Inconsistent Context.  & 64.4  & 70.2 & 70.2 & 72.5 \\
-w/o. Counterfactual Context. & 62.6 & 67.8 & 69.7 & 73.7 \\
\bottomrule
\end{tabular}}
\caption{Results (\%) of ablation study. 
EM/Acc in short-form tasks represents the average score between QA metrics (EM/Acc) and the accuracy of FaithEval.
N/A means a false generation pattern hacks this metric.}
\label{tb:ablation} 
\end{table}

\subsection{Analysis}

\noindent
\textbf{Ablation Study.}
We conduct an ablation study in Table \ref{tb:ablation}.
The result reveals that our proposed \textsc{Canoe} (including Dual-GRPO and the designed short-form data synthesis) significantly improves the faithfulness of LLMs in both short-form and long-form generation tasks.
For Dual-GRPO, we observe that directly applying GRPO to synthesized short-form data leads to over-optimizing short-form generation and a false response generation pattern.
We find that tuned models tend to directly copy text spans from the given context as the final answer instead of following instructions in long-form generation tasks (we show the case study in Table \ref{tab:sum_grpo_case} to visually show this phenomenon).
Thus, the generated responses do not contain syntactically and semantically complete sentences for long-form generation tasks, which leads to low QualityScore performance and also invalidates the metric used for evaluating faithfulness.
For the designed QA tasks used to ensure the complexity and diversity of training data, we replace the designed Reasoning-required Context and Inconsistent Context samples with the same number of samples that contain Straightforward Context.
We find that involving these more challenging instances can improve the effectiveness of RL training.
We also replace the data points that contain Counterfactual Context with the same number of factual samples.
The designed Counterfactual Context improves the final performance as it prevents models from depending on their learned factual knowledge to find the right answers.

\noindent
\textbf{Case Study.}
We find that directly applying GRPO instead of our Dual-GRPO to synthesized short-form data leads to over-optimizing short-form generation and a false response generation pattern.
As shown in Table \ref{tab:sum_grpo_case}, we find that the tuned model GRPO-LLaMA-8B tends to copy text spans from the given context as the final answer instead of following instructions in the summarization task.
We also observe the same phenomenon in the tasks of text simplification and long-form QA; more details can be found in the Appendix~\ref{appendix:grpo_vis}.
However, when we apply Dual-GPRO, we find that trained models can generate fluent and complete sentences.
Thus, Dual-GRPO not only improves the faithfulness of LLMs in two types of response generation but also ensures the utility.
We also conduct a case study in Table \ref{tab:case} to show the advantages of \textsc{Canoe}.
\textsc{Canoe} ensures the statements are faithful and comprehensive, and the text flows naturally.

\begin{table}[t]
    \scriptsize
    \centering
    \resizebox{\linewidth}{!}{
        \begin{tabular}{m{\hsize}}
        
        \specialrule{1pt}{1pt}{2pt}
        
        \sethlcolor{tabcolor3}\hl{\textbf{Text Summarization Task}} \\ 
        Inputs: \{Documents\} \\
        Instruction: Refer to the passage above and provide a summary as the response. \\
        
        \specialrule{0pt}{0pt}{1pt}
        \cdashline{1-1}[0.8pt/2pt]
        \specialrule{0pt}{0pt}{2pt}
        
        \sethlcolor{tabcolor6}\hl{\textbf{GRPO-LLaMA-8B}}: Comedy clubs. (\xmark) \\

        \specialrule{0pt}{0pt}{1pt}
        \cdashline{1-1}[0.8pt/2pt]
        \specialrule{0pt}{0pt}{2pt}
        \sethlcolor{tabcolor1}\hl{\textbf{\textsc{Canoe}-LLaMA-8B}}:
        Kareem Badr and his friends were able to turn around the struggling comedy club, the Hideout, by applying their skills in problem-solving and efficiency. They broke down the business into small chunks and figured out how to make each aspect better and more efficient. This approach helped them to improve the club's performance and eventually expand it. The passage also mentions that the growing popularity of stand-up comedy is fueled in part by younger audiences, who view humorous TV hosts as their primary source for news. Many comedy clubs now require talent, and there is a career and a living to be made by good comedians who are not superstars. The clubs also make money from selling drinks and dinners, as well as from cover charges and talent fees. In addition, digital media has given live comedy a boost, with one terrific joke being able to get a million hits online.
        (\cmark) \\
        \specialrule{1pt}{0.5pt}{0pt}   
        \end{tabular}}
\caption{Case study between the models trained via GRPO and our model for the summarization task.}
    
\label{tab:sum_grpo_case}
\end{table}

\begin{table}[t]
    \scriptsize
    \centering
    \resizebox{\linewidth}{!}{
        \begin{tabular}{m{\hsize}}
        \specialrule{1pt}{1pt}{2pt}
        \textbf{Question: What do Greeks say when they break plates?} \\
        
        \specialrule{1pt}{0pt}{1pt}
        \textbf{Qwen-2.5-Instruct-7B}:
        Based on the information provided in the passages, \sethlcolor{tabcolor5}\hl{there is no specific mention of what exactly Greeks say when they break plates during the plate-breaking ceremony.}  (\xmark) \\

        \specialrule{0pt}{0pt}{1pt}
        \cdashline{1-1}[0.8pt/2pt]
        \specialrule{0pt}{0pt}{2pt}
        \textbf{\textsc{Canoe}-Qwen-7B}:
        According to the information given, \sethlcolor{tabcolor4}\hl{in the Three's Company episode ``Opening Night,'' when Jack breaks the plates at the end of the celebration, he exclaims ``Opa!''.}(\cmark)
        \sethlcolor{tabcolor3}\hl{This suggests that ``Opa!'' might be a phrase commonly used in Greek culture during such festivities.} (\cmark)
        \sethlcolor{tabcolor1}\hl{``Opa!'' is often used to express joy, excitement, or approval, which aligns with the celebratory nature of plate smashing in Greek traditions.} (\cmark)  \\
        \specialrule{1pt}{0.5pt}{0pt}   
        \end{tabular}}
\caption{Case study from long-form QA task CLAPNQ. For different useful statements, we use different colors.}
    
\label{tab:case}
\end{table}

\noindent
\textbf{Human Evaluation.}
Evaluating long-form generation tasks remains challenging. Thus, we conduct human evaluation in the Appendix~\ref{appendix:human_eval} to show the effectiveness.
We find that our method reduces faithfulness hallucinations and also ensures the response quality for three long-form generation tasks.

\noindent
\textbf{Discussion.}
We discuss possible concerns in the Appendix~\ref{appendix:discussion}, e.g., the effect of the amount of synthesized data, multilingual transfer ability, and context length generalization of \textsc{Canoe}.

\section{Related Work}
Recently, the demand for utilizing LLMs to generate coherent text responses based on the provided contexts has continued to grow, particularly in text summarization and RAG scenarios.
However, LLMs are often criticized for generating outputs that deviate from the provided contents, namely \textit{faithfulness hallucination} \citep{li2022faithfulnessnaturallanguagegeneration, ji-survey,hit-survey}.
Many approaches have been proposed to improve the faithfulness of language models \citep{zhu-etal-2021-enhancing, jones2024teaching, hu2024mitigatinglargelanguagemodel,li2025hallucinationdilemmafactualityawarereinforcement,yang-etal-2025-longfaith, zhang-etal-2025-longreward,wang2025retrievalaugmentedgenerationconflictingevidence}.
The first line of work focuses on the inference stage, such as designing prompts to encourage context integration \citep{zhou-etal-2023-context}, improving context quality via explicit denoising \citep{xu2024recomp},  and context-aware decoding to amplify contexts \citep{shi-etal-2024-trusting}.
Although effective, these approaches primarily serve as a compensatory way rather than enabling the model to inherently learn to prevent generating unfaithful responses.
Thus, many studies attempt to apply post-training methods to improve the faithfulness.
\citet{bi2024contextdpoaligninglanguagemodels} utilizes constructed faithful and unfaithful short-form completions and applies DPO to align LLMs to be context-faithful in short-form QA tasks.
\citet{huang2025improvingcontextualfaithfulnesslarge} trains LLMs to discriminate between faithful and unfaithful responses in long-form QA tasks by unfaithful response synthesis and contrastive tuning.
\citet{li2025generatediscriminateevolveenhancing} introduces a self-evolving framework to maintain the faithfulness of LLMs in long-form QA tasks through iterative self-improvement.
\citet{duong2025scope} proposes a pipeline to generate a self-supervised task-specific dataset and applies preference training to enhance the faithfulness for a special task.
However, these methods struggle to consistently improve the faithfulness of LLMs across various tasks, as these methods are designed for specific tasks.
Thus, how to consistently improve the faithfulness of LLMs on different downstream tasks, including short-form and long-form generation tasks, still remains under-explored.

\section{Conclusion}
\label{subsection:conclusion}

In this paper, we propose \textsc{Canoe}, a systematic post-training method for teaching LLMs to remain faithful in both short-form and long-form generation tasks without human annotations. 
By synthesizing diverse short-form QA data and introducing Dual-GRPO, a tailored RL method with three well-designed rule-based rewards, \textsc{Canoe} effectively improves the faithfulness of LLMs.
We first synthesize short-form QA data with four diverse tasks to construct high-quality and easily verifiable training data without human annotation.
We then propose Dual-GRPO, a rule-based RL method that includes three tailored rule-based rewards derived from synthesized short-form QA data, while optimizing both short-form and long-form response generation simultaneously.
Experimental results show that \textsc{Canoe} consistently improves the faithfulness of LLMs across diverse downstream tasks.


\bibliography{custom}


\clearpage

\appendix
\section*{Appendix}

\noindent This appendix is organized as follows.  

\begin{itemize}  
    \item In Section~\ref{appendix:traning_data} \textit{Training Data Details}, we report the details of constructing training data, e.g., the used triples and introduction of four designed tasks.  
    
    \item In Section~\ref{appendix:grpo} \textit{Dual-GRPO Details}, we go into detail about the proposed Dual-GRPO, including the system prompt and formal expressions of three well-designed rewards.
    
    \item In Section~\ref{appendix:eval} \textit{Evaluation Details}, we show the details of evaluations, e.g., the introduction of the used benchmarks and evaluation prompts.
    
    \item In Section~\ref{appendix:implementation} \textit{Implementations Details}, we show the details of our implementation and training, e.g., hyperparameters.
    
    \item In Section~\ref{appendix:human_eval} \textit{Human Evaluation}, we show the implementation details of human evaluation.
    
    \item In Section~\ref{appendix:discussion} \textit{Discussion}, we discuss some possible questions about the proposed \textsc{Canoe}. For example, we discuss the effect of the amount of synthesized data for training.

\end{itemize}

\section{Training Data Details}
\label{appendix:traning_data}
\subsection{Triples from Wikidata}
To ensure the usability of the synthetic data and collected triples, we follow \citet{bi2024contextdpoaligninglanguagemodels} to collect entities corresponding to the top 1,000 most-visited Wikipedia pages from 2016 to 2023 and 41 relations selected by \citet{bi2024contextdpoaligninglanguagemodels} shown in Table \ref{tab:relations}.
The most-visited Wikipedia pages are based on monthly page views and retain the most popular entities using criteria such as the number of hyperlinks.
We finally collected 6,316 entities and 30,762 triples.
We randomly select these triples to synthesize our training data, and finally construct 10,000 samples as the final training data.

\subsection{Construction of Four Different Tasks}

We design four different tasks to enhance the complexity and diversity of our training data.
Meanwhile, we select \textit{GPT-4o-2024-08-06} to construct the contexts and questions.

\noindent
\textbf{Straightforward Context.}
As shown in Sec. \ref{sec:data}, we keep the original collected factual triple as input to query GPT-4o to synthesize the data $(c, q, a)$.
The prompts for querying GPT-4o to obtain the generated questions and contexts can be found in Figure \ref{fig:prompt_s_q} and Figure \ref{fig:prompt_s_c}.
We keep 2,000 such samples in the synthesized 10,000 training data, i.e., 20\% of the data.

\noindent
\textbf{Reasoning-required Context.}
We construct paths $[(h^1, r^1, t^1),...,(h^n, r^n, t^n)]_{n\leq4}$ from a sub-graph; more details can be found in Sec. \ref{sec:data}.
Then, we use the $n$-th tail entity $t^n$ as the ground truth answer and use the constructed paths to query GPT-4o to obtain the multi-hop context and question.
The prompts for querying GPT-4o to obtain the generated questions and contexts can be found in Figure \ref{fig:prompt_r_q} and Figure \ref{fig:prompt_r_c}.
We finally keep 2,000 such samples in the synthesized 10,000 training data, i.e., 20\% of the data.

\noindent
\textbf{Inconsistent Context.}
This involves multiple randomly ordered contexts generated from different triples.
This simulates noisy and inconsistent scenarios, where models need to detect inconsistencies and focus on useful and relevant contexts to answer the questions.
We construct such a sample by combining the contexts from up to three QA samples with reasoning-required context and use the original $t^n$ as the answer.
In this way, we can obtain more complex samples than ones with the reasoning-required context.
To avoid duplicating the 2,000 samples with the reasoning-required context collected above, we reconstruct the new samples with the reasoning-required context used to obtain the samples with the inconsistent context.
We keep 1,000 such samples in the synthesized 10,000 training data, i.e., 10\% of the data.

\noindent
\textbf{Counterfactual Context.}
A counterfactual context includes statements that go against common sense found in the collected triples.
Specifically, we construct samples with counterfactual contexts below by modifying previously collected triples (of three types, including straightforward context, reasoning-required context, and inconsistent context).
We replace the tail entity $t$ of the original collected triple with a similar but counterfactual entity  $t^{cf}$, which is obtained by query GPT-4o using prompt \textit{``Generate me a noun for an entity that is similar to the \{$t$\} but different, and require the entity to exist in the real-world, please tell me the answer directly:''}. 
Then, we query GPT-4o to generate questions and counterfactual contexts to construct counterfactual samples, using the counterfactual triples.
The prompts used in constructing samples with counterfactual contexts are the same as the prompts used in constructing the three different tasks above.
The reason we construct samples with counterfactual context in this way is that this prevents the model from learning the appropriate factual knowledge to answer the question correctly, rather than correctly exploiting the given contextual information.
Therefore, we construct the same number of samples as the summed number of the three types above (including straightforward context, reasoning-required context, and inconsistent context), i.e., 5000 samples (50\% of the data).
This task stresses the importance of keeping answers faithful in contexts, as it stops them from relying solely on the learned knowledge of LLMs to provide correct answers.

\begin{table}
\centering
\footnotesize
\arrayrulecolor{black}
\begin{tabular}{lccc} 
\toprule
\multicolumn{1}{l}{Type} & Num & Avg Len \\
\cmidrule(lr){1-4}
{Straightforward Context.}  & 2,000 & 186.3 \\
{Reasoning-required Context.} & 2,000 & 262.2 \\
{Inconsistent Context.}  & 1,000  & 421.2 \\
{Counterfactual Context.} & 5,000 & 260.8 \\
\toprule
\end{tabular}
\caption{Statistics of the training data.
Num shows the number of samples.
Avg Len shows the average length of the samples.}
\arrayrulecolor{black}
\label{tab_training_data_statistics}
\end{table}

\subsection{Statistics}
We show the statistics of the training data in Table \ref{tab_training_data_statistics}.
Even though the length of the data we synthesize is short, we find that our model can be generalized with consistently state-of-the-art results on a wide range of tasks with different input lengths by utilizing our proposed Dual-GRPO, e.g., long-form QA and RAG generation with long texts as inputs.

\section{Dual-GRPO Details}
\label{appendix:grpo}

In this section, we give a more detailed introduction to our proposed Dual-GRPO, including the designed system prompt and formal expressions of three different rewards. 

\noindent
\textbf{System Prompt.}
For the provided contextual information and question, Dual-GRPO employs the designed system prompt that requires LLMs to produce a reasoning process, then a long-form answer that consists of detailed and complete sentences, and finally a concise short-form answer in just a few words.
In this way, we can assign different reward scores to long-form answers and short-form answers while optimizing them both at once.
Meanwhile, this system prompt also triggers zero-shot chain-of-thought reasoning in the policy model, which progressively improves as training advances to optimize for the reward.
We use the same system prompt to train both LLaMA and Qwen models.
We show our used system prompt in Figure \ref{fig:sys_prompt}.

\noindent
\textbf{Accuracy Reward.}
For short-form generation, we directly assign the accuracy reward. 
Specifically, for the generated short-form response $y_{sf}$ based on the given context $c$ and question $q$, which is extracted from the whole generated response $y_{whole}$ via string matching, and the ground truth answer $y_{gt}$ from the synthesized training data, the accuracy reward $R_{\text{acc}}$ for the LLM $\theta$ can be calculated as:
$$
R_{\text{acc}} = 
\begin{cases} 
1 & \text{if } y_{sf}(c,q | \theta) =y_{gt}, \\ 
0 & \text{otherwise}.
\end{cases}
$$

We use the exact matching (EM) to measure accuracy, giving a score of 1 for a match and 0 for a mismatch. 
Thus, we can ensure that the generated short-form response correctly answers the question based on the given context, making LLMs more faithful in short-form response generation.

\noindent
\textbf{Proxy Reward.}
Instead of directly evaluating the faithfulness of the generated long-form response, we propose a proxy reward to evaluate it implicitly.
For each generated long-form answer $y_{lf}$, we replace the given context $c$ with it as new input and infer the LLM $\theta$ to determine whether the LLM can produce the correct short-form answer $y_{sf}$ based on $y_{lf}$ for the question $q$.
Proxy reward $R_{\text{proxy}}$ can be calculated as:
$$
R_{\text{proxy}} = 
\begin{cases} 
1 & \text{if } y_{sf}(y_{lf},q| \theta) =y_{gt}, \\ 
0 & \text{otherwise}.
\end{cases}
$$

If the generated long-form response can help LLMs generate the correct answer, it indicates that the long-form response is faithful to the context, contains meaningful sentences, and correctly addresses the question.
Thus, we assign a reward score of 1 for the positive long-form response that helps the LLM to produce the correct answer, and a reward score of 0 for those that lead to incorrect answers.

\noindent
\textbf{Format Reward.}
To enforce the desired output format, we assign a reward on the whole generated response $y_{whole}$ to evaluate whether it contains the proper XML tags.
We use three types of tags as shown in our system prompt, as shown in Figure \ref{fig:sys_prompt}, including \textless think\textgreater, \textless long\_answer\textgreater, and \textless short\_answer\textgreater tags.
Formally,
$$
R_{\text{format}} = 
\begin{cases} 
1 & \text{if correct formatting is present,} \\ 
0 & \text{if incorrect formatting}.
\end{cases}
$$

We use the string matching method to evaluate whether the responses adhere to the format.

\noindent
\textbf{Final Reward.}
Finally, we use the sum of these three rewards as the final composite reward $R_{\text{final}}$.
This well-designed reward $R_{\text{final}}$ of Dual-GRPO enhances the efficacy of the rule-based RL training framework to guide the model toward generating more faithful responses in both short-form and long-form tasks.
Formally,
$$
R_{\text{final}} = R_{\text{acc}} + R_{\text{proxy}} + R_{\text{format}}.
$$

Finally, we use this reward $R_{\text{final}}$ to compute an advantage $A_i$ for each output, guiding policy updates according to the rule-based GRPO objective.

\noindent
\textbf{Potential Reward Hacking Concerns.}
In the early experiments, we have also tried adding the length reward for long-form responses (i.e., the content between \textless long\_answer\textgreater and \textless /long\_answer\textgreater tags) to avoid the potential reward hacking, e.g., avoiding the policy model directly copying the given context as the long-form response, but found that the task performance does not have a significant difference.

\section{Evaluation Details}
\label{appendix:eval}

\subsection{Datasets}

\noindent
\textbf{ConFiQA (Counterfactual QA).}
This is a dataset that incorporates knowledge conflicts through counterfactual passages to evaluate the faithfulness of LLMs on short-form generation.
ConFiQA consists of three tasks: QA (Question Answering), MR (Multi-hop Reasoning), and MC (Multi-Conflicts). 
QA features single-hop question-answering tasks with context containing one corresponding counterfactual, while MR and MC involve multi-hop reasoning tasks with context containing one and multiple related counterfactual contexts, respectively.
ConFiQA contains 1,500 data points used for testing (500/500/500 from QA/MC/MR).

\noindent
\textbf{CNQ (Counterfactual QA).}
CNQ is constructed based on Natural Questions \citep{kwiatkowski-etal-2019-natural}.
In CNQ, the context is modified to support counterfactual answers following \citep{longpre-etal-2021-entity}.
It contains 2,773 samples that incorporate counterfactual passages to evaluate the faithfulness of LLMs on short-form generation.

\noindent
\textbf{FaithEval (Counterfactual Multiple-choice QA).}
FaithEval is a novel and comprehensive benchmark tailored to evaluate the faithfulness of LLMs in contextual scenarios across three diverse tasks: unanswerable, inconsistent, and counterfactual contexts.
We select the counterfactual task to evaluate the faithfulness, which contains 1,000 multiple-choice QA samples curated based on ARC-Challenge \citep{clark2018thinksolvedquestionanswering}.

\noindent
\textbf{FiQA (Factual QA).}
FiQA is a factual version of ConFiQA, which shares the same questions as ConFiQA but contains the factual contexts and answers.
The contexts and answers are provided by \citet{bi2024contextdpoaligninglanguagemodels}, thus we can evaluate the faithfulness of LLMs in factual short-form response generation.
It contains 1,500 samples for evaluation.

\noindent
\textbf{FollowRAG (RAG Scenarios for short-form QA).}
FollowRAG aims to assess the model’s ability to follow user instructions in complex multi-document contexts.
It consists of four well-known open-domain QA datasets for RAG scenarios, including NaturalQA, TriviaQA, HotpotQA, and WebQSP.
We utilize the provided passages in FollowRAG as context and original query (instead of the version with added instruction constraints proposed by \citet{dong2024generalinstructionfollowingalignmentretrievalaugmented}) as questions.
We also use the original answers to report the results.
FollowRAG contains 2,800 samples used for testing (700/700/700/700 from NaturalQA/TriviaQA/HotpotQA/WebQSP).
Different from short-form generation tasks that the contexts always contain answers, in real-world RAG scenarios, the answer may not appear in the retrieved passages, and these passages tend to be noisy.

\noindent
\textbf{XSum (Summarization).}
Text summarization is a content-grounded task where a model is provided a piece of text and tasked with synthesizing the most salient information within that text.
XSum is a widely used dataset for text summarization, which consists of about 220,000 BBC articles as input documents. 
To facilitate our evaluation, we use the first 1,000 data points from the test set to evaluate our method.

\noindent
\textbf{WikiLarge (Simplification).}
Simplification is a task where a model is provided a piece of text and is tasked with paraphrasing it to make the text easier to read and understand.
We use 1k instances sampled from the WikiLarge dataset as a test set, following \citet{ravichander2025halogenfantasticllmhallucinations}.

\noindent
\textbf{CLAPNQ (Long-form QA).}
CLAPNQ is a grounded long-form QA benchmark dataset for Retrieval Augmented Generation of LLMs.
The answers are typically long, 2-3 sentences grounded on a single gold passage, in contrast to datasets based on machine reading comprehension, such as short-form Natural Questions, which are just a few words.
CLAPNQ includes long answers with grounded gold passages from Natural Questions.
We utilize the provided passages and questions from the dev set to evaluate the faithfulness of LLMs in long-form response generation for open-domain questions, which contains 600 data points.

\subsection{Metrics and LLM-as-a-Judge}
\label{appendix:llm-as-judge}
\noindent
\textbf{Metrics for Short-form Generation Tasks.}
We evaluate performance based on whether gold answers are included in the generated responses (i.e., Acc) following \citet{asai2024selfrag} and exact matching (EM) for QA tasks.
For multiple-choice questions in FaithEval, we use keyword matching to verify the accuracy, i.e., Acc.

\noindent
\textbf{Metrics for Long-form Generation Tasks.}
To evaluate the faithfulness in long-form generation tasks, we use MiniCheck to check whether the model response is grounded in the provided context.
MiniCheck is the SOTA method to recognize if LLM output can be grounded in given contexts. 
We select the MiniCheck-FT5\footnote{https://huggingface.co/lytang/MiniCheck-Flan-T5-Large} because it is the best fact-checking model, outperforming GPT-4o in evaluating the faithfulness.
If the model response contains at least one statement that cannot be inferred from the context, we consider it as a negative response; otherwise, it is a positive response.
To evaluate the quality of the generated long-form responses for three different tasks (QualityScore), including summarization, simplification, and long-form QA, we design different prompts to query \textit{GPT-4o-2024-11-20} as a judge to get the quality scores.
We report the average results of the quality score results by querying GPT-4o twice.
The prompts for three tasks can be found in Figure \ref{fig:prompt_scoring_sum}, Figure \ref{fig:prompt_scoring_sim}, and Figure \ref{fig:prompt_scoring_lfqa}.

\subsection{Baselines}
For SOTA LLMs, we select the following versions of these models to report the results.
Specifically, we use \textit{GPT-4o-2024-08-06} for GPT-4o, \textit{GPT-4o-mini-2024-07-18} for GPT-4o-mini, \textit{Claude 3.7 Sonnet-2025-02-19} for Claude 3.7 Sonnet and Claude 3.7 Sonnet-thinking, \textit{Deepseek R1 2025-01-20} for Deepseek R1, \textit{Deepseek V3 2024-12-26} for Deepseek V3, and \textit{o1-2024-12-17} for OpenAI o1.
To get stable experimental results, we query these models twice and report the average results on each task.
For the methods that are designed for improving the faithfulness, we reproduce their released code based on \textit{LLaMA-3-Instruct} and \textit{Qwen-2.5-Instruct}.
For SCOPE, we train it on the 10,000 sampled training set of the summarization task XSum as SCOPE$_{sum}$, which keeps the same number of data we used for training \textsc{Canoe} and provides a fair comparison.

\begin{table*}[h]
\scriptsize
\centering  
\resizebox{0.83\textwidth}{!}{
\begin{tabular}{lccccccccc}
\toprule
\multicolumn{1}{l}{\multirow{2}{*}{\textbf{Model}}}
& \multicolumn{2}{c}{\textbf{HotpotQA}} & \multicolumn{2}{c}{\textbf{NaturalQA}} & \multicolumn{2}{c}{\textbf{TriviaQA}} & \multicolumn{2}{c}{\textbf{WebQSP}}  \\
\quad &  \textbf{EM} & \textbf{Acc} & \textbf{EM}   & \textbf{Acc} &\textbf{EM} & \textbf{Acc} &\textbf{EM} & \textbf{Acc} \\
\midrule
\rowcolor{mygray} \multicolumn{9}{c}{\cellcolor{myyellow} \textbf{The state-of-the-art LLMs}} \\
GPT-4o & 24.7 &	32.0 	&37.0 &	55.0 	&62.3 &	72.3 &	44.9 &	71.7  \\
GPT-4o mini & 18.0 &	26.2 &	35.0 	&48.2 &	59.5 &	65.5 &	41.4 &	65.3  \\
DeepSeek V3 & 18.7 &	27.7 &	34.9 	&54.3 &	60.0 &	70.0 &	37.1 &	68.9  \\
Claude 3.7 Sonnet &15.3 	&24.1 	&33.6 	&53.9 &	62.5 	&72.5 &	33.7 &	64.3  \\
\hdashline[2pt/3pt]
OpenAI o1 & 27.0 &	34.0 &	37.0 	&50.0 &	63.0 	&76.0 &	35.0 	&68.0  \\
DeepSeek R1 & 26.0 	&29.3 &	38.7 	&52.9 &	68.0 &	73.0 &	38.9 	&71.3  \\
Claude 3.7 Sonnet-Thinking  \quad \quad \quad \quad \quad \quad \quad \quad \quad \quad \quad \quad \quad \quad  \quad \quad & 20.1 	&30.2 	&35.6 	&53.0 &	63.4 &	72.0 &	36.0 &	66.0 \\
\midrule
\rowcolor{mygray} \multicolumn{9}{c}{\cellcolor{myyellow} \textbf{LLaMA-3-Instruct Series}} \\
LLaMA-3-Instruct-8B  &13.0 	&18.2 &	31.0 &	40.3& 	45.5 &	60.2 	&35.0 &	60.4  \\
LLaMA-3-Instruct-70B & 24.1 &	28.7 &	36.5 &	45.3 &	63.0 &	66.6 &	31.3 &	42.1  \\
\hdashline[2pt/3pt]
SFT-8B&  3.7 &	5.4 &	15.9 &	18.7 &	26.6 &	26.3 &	30.4& 	33.6  \\
Context-DPO-8B & 10.1 &	16.7 &	23.4 &	37.8 &	53.3 &	62.3 &	32.8 	&58.3  \\
SCOPE$_{sum}$-8B &12.0 &	20.5 &	25.7 	&42.5 	&46.4 &	58.6 &	36.1 	&63.2  \\
\rowcolor{blue!5} \textbf{\textsc{Canoe}-LLaMA-8B} & \textbf{21.4} &	\textbf{23.3} &	\textbf{37.4} &	\textbf{46.9} &	\textbf{60.0} &	\textbf{67.3} &	\textbf{44.9} 	& \textbf{69.3}  \\
\midrule
\rowcolor{mygray} \multicolumn{9}{c}{\cellcolor{myyellow} \textbf{Qwen-2.5-Instruct Series}} \\
Qwen-2.5-Instruct-7B & 14.0 &	17.6 &	32.2 &	42.3 &	50.3 	&62.3& 	33.9 &	58.8  \\
Qwen-2.5-Instruct-14B &17.5 &	21.7 &	29.3 &	48.0 &	55.6& 	69.3 	&36.9& 	65.7  \\
Qwen-2.5-Instruct-32B & 16.5 &	24.6 &	26.3& 	50.2 &	50.0 &	70.7 &	42.7 &	66.7  \\
Qwen-2.5-Instruct-72B & 21.8 	&28.0 &	34.5 &	51.0 	&61.8 &	73.0 &	35.7 &	70.6  \\
\hdashline[2pt/3pt]
SFT-7B &16.2 &	18.3 	&26.5 	&30.2 &	43.2 &	58.2 	&30.2 	&60.2  \\
Context-DPO-7B  &13.0 &	17.2 	&25.2 	&40.2 	&50.1& 	63.2 &	35.7 &	54.3  \\
SCOPE$_{sum}$-7B &12.5 &	19.5 &	27.2 &	43.5 &	48.4 &	60.1 &	34.2 &	60.7  \\
\rowcolor{blue!5}  \textbf{\textsc{Canoe}-Qwen-7B} &\textbf{18.0} 	&\textbf{22.6}& 	\textbf{35.7} &	\textbf{47.4} &	\textbf{57.4} 	&\textbf{65.7} &	\textbf{36.9}& 	\textbf{65.0} \\
\hdashline[2pt/3pt]
\rowcolor{blue!5}  \textbf{\textsc{Canoe}-Qwen-14B} & \textbf{19.9} &	\textbf{25.7} &	\textbf{41.9} &	\textbf{51.6} &	\textbf{63.3} &	\textbf{71.7} &	\textbf{59.4} &	\textbf{69.3}  \\

\bottomrule
\end{tabular}
}
\caption{Experimental results (\%) on FollowRAG. \textbf{Bold} numbers indicate the best performance of models with the same size.}
\label{tb:follow_rag}
\end{table*}

\subsection{More Detailed Experimental Results}
FollowRAG contains four different QA datasets in RAG scenarios. 
We report the average results in Table \ref{tb:main}.
We show the more detailed results of FollowRAG in Table \ref{tb:follow_rag}.

\subsection{Test-time Prompts}
\label{appendix:test-time-prompt}
For baselines, the prompts for different tasks can be found in Figure \ref{fig:prompt_sfqa}, Figure \ref{fig:prompt_mcqa}, Figure \ref{fig:prompt_sum}, Figure \ref{fig:prompt_sim}, and Figure \ref{fig:prompt_lfqa}.
It is worth noting that some test-time prompt templates differ from those in the original paper, which may lead to some variation in the reported results.
For example, in the original paper of Context-DPO \citep{bi2024contextdpoaligninglanguagemodels}, the authors used a very simple prompt template, i.e., \{Passages\} Q: \{\} A:, to report performance results of popular LLMs on the ConFiQA dataset. 
Such a simple prompt template will lead to a very low EM score when the grounded-truth answer is just an entity, which may lead to unfair comparison and useless evaluation. 
Therefore, we redesign the corresponding prompt templates by adding length constraints as shown in Figure \ref{fig:prompt_sfqa} to ensure a fair comparison.
Meanwhile, since the authors of FaithEval \citep{ming2025faitheval} do not provide official evaluation scripts and a specific task prompt for FaithEval-Counterfactual, we write our evaluation scripts and use the prompt shown in Figure \ref{fig:prompt_mcqa} in our experiments.
We find that the state-of-the-art LLMs, e.g., GPT-4o, show stable performance consistent with the original papers.
At the same time, we observe that slightly smaller models, such as LLaMA-3-Instruct-8B, show decreased performance on our designed prompt templates.
To evaluate the factuality of LLMs, we modify the original FaithEval and make it a closed-book QA setting, and use the prompts shown in Figure \ref{fig:prompt_faitheval_fact}.

During the evaluation for \textsc{Canoe}, we apply the same system prompt during the Dual-GRPO training, and extract the content between \textless short\_answer\textgreater and \textless/short\_answer\textgreater tags as the final answers for short-form generation tasks.
Also, for long-form generation tasks, we extract the content between \textless long\_answer\textgreater and \textless /long\_answer\textgreater tags as the final answers.
We also find that the long-form responses generated by \textsc{Canoe} can provide correct answers in short-form generation tasks in the Appendix \ref{appendix:can_long}.
Thus, for real-world applications, we recommend using the generated long-form responses as the system responses for the user’s instructions, because these long-form responses can not only faithfully complete long-form generation tasks, but also provide correct answers in short-form generation tasks.

\begin{figure}[t]
    \centering
    \includegraphics[width=7cm]{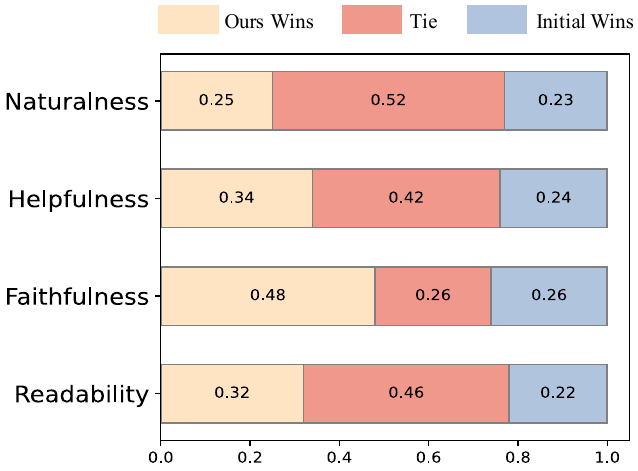}
    \caption{Human evaluation across four key dimensions. }
    \label{fig_human}
\end{figure}

\section{Implementation Details}
\label{appendix:implementation}
We implement our method based on the RL framework open-r1 \citep{openr1}.
We use AdamW optimizer \citep{ loshchilov2019decoupledweightdecayregularization}, with a $1\times10^{-6}$ learning rate, a batch size of 14 for 7B/8B models, and a batch size of 7 for the 14B model, steering the training across two epochs.
We set the maximum input length for the models to 1,024 and the maximum generation length to 1,024. 
The number of generations $G$ during the RL training is set to 7, which is used in Eq. (\ref{grpo_1}).
We set 0.04 for $\beta$ used in Eq. (\ref{grpo_1}).
We set 0.2 for $\epsilon$ used for the clip shown in Eq. (\ref{grpo_2}).
We set 0.9 for temperature in RL training to generate responses.
We conduct our experiments on NVIDIA A800-80G GPUs with DeepSpeed+ZeRO2 for 7B/8B models, DeepSpeed+ZeRO2+Offloading for the 14B model, and BF16.
During the inference, we set 0.7 for temperature for the evaluation of our models and baselines. 
For each task, we infer twice and report the average scores as final results.

\begin{table*}[!h]
\scriptsize
\centering  
\resizebox{0.95\textwidth}{!}{
\begin{tabular}{lcccccccccc}
\toprule
\multicolumn{1}{l}{\multirow{2}{*}{\textbf{Model}}}
& \multicolumn{1}{c}{\textbf{ConFiQA}} & \multicolumn{1}{c}{\textbf{FiQA}} & \multicolumn{1}{c}{\textbf{CNQ}}&  \multicolumn{1}{c}{\textbf{FaithEval}} & \multicolumn{1}{c}{\textbf{HotpotQA}} & \multicolumn{1}{c}{\textbf{NaturalQA}} & \multicolumn{1}{c}{\textbf{TriviaQA}} & \multicolumn{1}{c}{\textbf{WebQSP}} & \multicolumn{1}{c}{\multirow{2}{*}{\textbf{Avg}}}\\
\quad &  \textbf{Acc} & \textbf{Acc} & \textbf{Acc}   & \textbf{Acc} &\textbf{Acc} & \textbf{Acc} &\textbf{Acc} & \textbf{Acc} \\
\midrule
\rowcolor{mygray} \multicolumn{10}{c}{\cellcolor{myyellow} \textbf{The state-of-the-art LLMs}} \\
GPT-4o & 42.7 &	79.6 	&55.9 &	47.5& 	32.0 &	55.0 &	72.3 &	71.7 	&57.1  \\
GPT-4o mini  &63.7 	&78.8& 	54.3 &	50.9 &	26.2 &	48.2 &	65.5 &	65.3 &	56.6  \\
DeepSeek V3 & 58.6& 	76.5 &	67.3 	&51.0 &	27.7 &	54.3 	&70.0 &	68.9 	&59.3 \\
Claude 3.7 Sonnet & 36.0 	&72.2 &	65.0 &	45.6 	&24.1 &	53.9 &	72.5 &	64.3 &	54.2 \\
\hdashline[2pt/3pt]
OpenAI o1 & 57.9 &	89.7 &	39.1 	&52.0& 	\textbf{34.0} 	&50.0 &	\textbf{76.0} 	&68.0 &	58.3  \\
DeepSeek R1& 74.3 &	80.7 &	70.2 &	60.1 &	29.3 &	52.9 &	73.0 &	71.3 &	64.0  \\
Claude 3.7 Sonnet-Thinking&  38.7 &	76.7 &	67.0 &	57.0 &	30.2 &	53.0 &	72.0 &	66.0 	&57.6  \\
\midrule
\rowcolor{mygray} \multicolumn{10}{c}{\cellcolor{myyellow} \textbf{LLaMA-3-Instruct Series}} \\
LLaMA-3-Instruct-8B & 58.2& 	59.3 &	45.2 	&52.0 &	18.2 &	40.3& 	60.2 &	60.4 	&49.2  \\
LLaMA-3-Instruct-70B &  54.5 &	66.8 	&65.0 &	50.9 &	28.7& 	45.3 &	66.6& 	42.1 &	52.5  \\
\hdashline[2pt/3pt]
SFT-8B&  70.3& 	59.9 	&65.7 &	43.0 	&5.4 &	18.7 &	26.3 &	33.6 &	40.4  \\
Context-DPO-8B&  72.9 &	59.5 &	62.3 &	37.5 &	16.7 &	37.8 &	62.3 	&58.3 &	50.9  \\
SCOPE$_{sum}$-8B& 64.6 &	68.7 &	60.6 &	55.7 &	20.5 &	42.5 	&58.6& 	63.2 &	54.3  \\
\rowcolor{blue!5} \textbf{\textsc{Canoe}-LLaMA-8B}  &80.9 &	84.9 &	73.4 &	74.6 &	23.3 	&46.9 &	67.3 &	69.3 	&65.1  \\
\rowcolor{blue!5} \textbf{\quad - Using Generated Long-form Responses.} & \textbf{92.3} &	\textbf{95.5} &	81.6 &	\textbf{78.2} &	32.7& 	\textbf{59.3} &	74.1& 	\textbf{79.1} &	\textbf{74.1} \\
\rowcolor{blue!5}  \quad $\Delta$ Compared to Using Generated Short-from Response.  & \textcolor[rgb]{0.7,0,0}{+11.4} 	& \textcolor[rgb]{0.7,0,0}{+10.6} 	& \textcolor[rgb]{0.7,0,0}{+8.2} 	& \textcolor[rgb]{0.7,0,0}{+3.6} 	& \textcolor[rgb]{0.7,0,0}{+9.4} 	& \textcolor[rgb]{0.7,0,0}{+12.4} 	& \textcolor[rgb]{0.7,0,0}{+6.8} 	& \textcolor[rgb]{0.7,0,0}{+9.8} 	& \textcolor[rgb]{0.7,0,0}{+9.0} \\ 
\midrule
\rowcolor{mygray} \multicolumn{10}{c}{\cellcolor{myyellow} \textbf{Qwen-2.5-Instruct Series}} \\
Qwen-2.5-Instruct-7B& 61.0 &	68.4 	&68.2 &	56.1 &	17.6 	&42.3& 	62.3 &	58.8 &	54.3  \\
Qwen-2.5-Instruct-14B &47.3 &	61.4 &	64.3 &	51.6 &	21.7& 	48.0 &	69.3 	&65.7 &	53.7  \\
Qwen-2.5-Instruct-32B  &66.4 	&81.1 &	66.4 &	47.0 &	24.6 	&50.2 &	70.7 &	66.7 &	59.1  \\
Qwen-2.5-Instruct-72B & 52.3 &	67.3 &	62.2 &	45.2 &	28.0 &	51.0 &	73.0 &	70.6 &	56.2  \\
\hdashline[2pt/3pt]
SFT-7B & 69.8 &	76.6 	&65.3 &	50.3 	&18.3 &	30.2 &	58.2 &	60.2 &	53.6  \\
Context-DPO-7B & 70.6 &	78.2 &	70.1 &	45.7 &	17.2 &	40.2 &	63.2 	&54.3 &	54.9   \\
SCOPE$_{sum}$-7B&  47.9 &	60.9 &	55.3 &	52.3 &	19.5 &	43.5 	&60.1 &	60.7 &	50.0  \\
\rowcolor{blue!5}  \textbf{\textsc{Canoe}-Qwen-7B} &75.2 	&83.5 	&76.4 &	70.5 &	22.6 &	47.4 &	65.7 &	65.0 &	63.3  \\
\rowcolor{blue!5}  \textbf{\quad - Using Generated Long-form Responses.} &82.9 &	92.3 &	83.2 &	73.2 &	29.8 &	56.9 	&70.6& 	72.7 &	70.2  \\
\rowcolor{blue!5}  \quad $\Delta$ Compared to Using Generated Short-from Response.  & \textcolor[rgb]{0.7,0,0}{+7.7} 	& \textcolor[rgb]{0.7,0,0}{+8.8} 	& \textcolor[rgb]{0.7,0,0}{+6.8} 	& \textcolor[rgb]{0.7,0,0}{+2.7} 	& \textcolor[rgb]{0.7,0,0}{+7.2} 	& \textcolor[rgb]{0.7,0,0}{+9.5} 	& \textcolor[rgb]{0.7,0,0}{+4.9} 	& \textcolor[rgb]{0.7,0,0}{+7.7} 	& \textcolor[rgb]{0.7,0,0}{+6.9}  \\
\hdashline[2pt/3pt]
\rowcolor{blue!5}  \textbf{\textsc{Canoe}-Qwen-14B}  &87.4 &	88.5 &	84.2 &	67.4 &	25.7 &	51.6 &	71.7 &	69.3 &	68.2  \\
\rowcolor{blue!5}  \textbf{\quad - Using Generated Long-form Responses.} & 89.8 &	94.4 &	\textbf{87.1} &	70.6 &	30.0 &	58.0 &	73.1 &	76.6 &	72.5 \\
\rowcolor{blue!5}  \quad $\Delta$ Compared to Using Generated Short-from Response.  & \textcolor[rgb]{0.7,0,0}{+2.4} 	& \textcolor[rgb]{0.7,0,0}{+5.9} 	& \textcolor[rgb]{0.7,0,0}{+2.9} 	& \textcolor[rgb]{0.7,0,0}{+3.2} 	& \textcolor[rgb]{0.7,0,0}{+4.3} 	& \textcolor[rgb]{0.7,0,0}{+6.4} 	& \textcolor[rgb]{0.7,0,0}{+1.4} 	& \textcolor[rgb]{0.7,0,0}{+7.3} 	& \textcolor[rgb]{0.7,0,0}{+4.2}  \\
\bottomrule
\end{tabular}
}
\caption{Experimental accuracy score results (\%) on short-form generation tasks. 
\textbf{Bold} numbers indicate the best performance among all the models.}
\label{tb:acc_long_form}
\end{table*}

\section{Human Evaluation}
\label{appendix:human_eval}
Evaluating long-form generation tasks remains challenging.
Thus, we conduct a human evaluation on the 90 samples from long-form generation tasks, including 30/30/30 for summarization/simplification/long-form QA tasks.
We evaluate these samples across four key dimensions: readability, faithfulness, helpfulness, and naturalness.
For each comparison, three options are given (Ours Wins, Tie, and Initial Model Wins), and the majority voting determines the final result. 
The participants follow the principles in Figure \ref{fig:human_evaluation_principles} to make the decision.
We invite three Ph.D. students to compare the responses generated by the models. 
Before participants begin to make judgments, we describe the principles of our design in detail and ensure that each participant correctly understands the principles. 
If the final result can not be determined by majority voting, we will hold a discussion among the participants and vote on the result again.
We compare two models, including \textsc{Canoe}-LLaMA-8B as our method and LLaMA-3-Instruct-8B as the initial model.
As shown in Figure \ref{fig_human}, we can find that \textsc{Canoe} reduces faithfulness hallucinations and also ensures the response quality for three long-form generation tasks.

\section{Discussion}
\label{appendix:discussion}

\textbf{Can Long-form Responses Generated by \textsc{Canoe} Provide Correct Answers in Short-form Generation Tasks?}
\label{appendix:can_long}
This exploration is important because, in real-world applications, it is difficult to pre-determine whether to use generated short-form responses (i.e., the context between \textless short\_answer\textgreater and \textless /short\_answer\textgreater tags) or long-form responses (i.e., the context between \textless long\_answer\textgreater and \textless /long\_answer\textgreater tags) as answers to respond to user instructions.
This contrasts with the evaluation of LLMs on different datasets, as described in the test-time strategies outlined in \ref{appendix:test-time-prompt}.
Thus, we first explore whether the long-form responses generated by \textsc{Canoe} (i.e., the context between \textless long\_answer\textgreater and \textless /long\_answer\textgreater tags) can provide correct answers in short-form generation tasks.
As shown in Table \ref{tb:acc_long_form}, when evaluating the generated long-form responses that contain the free-form answers, the accuracy scores consistently increase in all the short-form generation tasks compared to using the generated short-form responses.
It indicates that the generated short-form responses maintain conciseness, which is important for measuring the EM score, but can slightly reduce the accuracy score.
Therefore, in real-world applications, we can directly use the generated long-form responses as the system responses for the user’s instructions, because these long-form responses can not only efficiently and faithfully complete long-form generation tasks, but also provide correct answers in short-form generation tasks.

\begin{table}
\centering
\footnotesize
\arrayrulecolor{black}
\resizebox{0.97\linewidth}{!}{
\begin{tabular}{lccc} 
\toprule
\multicolumn{1}{l}{\textbf{Model}} & \textbf{Accuracy} & \textbf{Proxy} & \textbf{Format} \\
\cmidrule(lr){1-4}
\textbf{\textsc{Canoe}-LLaMA-8B}  & 70.3 & 66.1 & 99.4 \\
\textbf{\textsc{Canoe}-Qwen-7B} & 64.1 & 63.4  & 99.9 \\
\textbf{\textsc{Canoe}-Qwen-14B} &83.5 &  76.5 & 100.0 \\
\toprule
\end{tabular}}
\caption{Final rewards (\%) in the RL training stage.}
\arrayrulecolor{black}
\label{tab_rewards}
\end{table}

\noindent
\textbf{Final Rewards in the RL Training Stage.}
We show the final rewards in Table \ref{tab_rewards}.
We can find that models can easily learn the designed format, while accuracy and proxy rewards still remain challenging.
Meanwhile, in the early stages of RL training, the format reward increases quickly and converges rapidly, and as training proceeds, the accuracy reward and the proxy reward gradually increase and eventually converge.
This indicates that our well-designed training data construction strategy is effective and ensures the complexity and diversity, avoiding overfitting and reward hacking.

\begin{table}[!t]
\scriptsize	
\centering
\resizebox{0.97\linewidth}{!}{
\begin{tabular}{lccc}
\toprule
\multicolumn{1}{l}{\multirow{1}{*}{\textbf{Model}}} & \multicolumn{1}{c}{\textbf{MultiFieldQA-zh}} & \multicolumn{1}{c}{\textbf{DuReader}} & \multicolumn{1}{c}{\textbf{VCSUM}}  \\
\midrule
LLaMA-3-Instruct-8B  & {80.1} & 65.2 & 42.2 \\
\rowcolor{blue!5} \textbf{\textsc{Canoe}-LLaMA-8B} & \textbf{88.2} & \textbf{75.3} & \textbf{65.2} \\
\hdashline[2pt/3pt]
Qwen-2.5-Instruct-7B  & 82.3 & 70.3 & 45.5 \\
Qwen-2.5-Instruct-14B  & 83.5 & 72.2 & 47.8 \\
Qwen-2.5-Instruct-32B  & 85.1 & 77.2 & 52.7 \\
Qwen-2.5-Instruct-72B  & 88.9 & 80.1 & 57.1  \\
\hdashline[2pt/3pt]
\rowcolor{blue!5} \textbf{\textsc{Canoe}-Qwen-7B} & \textbf{90.1} & \textbf{78.3} & \textbf{66.5} \\
\rowcolor{blue!5} \textbf{\textsc{Canoe}-Qwen-14B} & \textbf{93.2} & \textbf{84.3} & \textbf{70.4} \\
\bottomrule
\end{tabular}}
\caption{Results (\%) on three Chinese datasets.
\textbf{Bold} numbers indicate the best performance of models with the same model size.}
\label{tab_chinese}
\end{table}

\begin{figure}[!t]
    \centering
    \includegraphics[width=6.8cm]{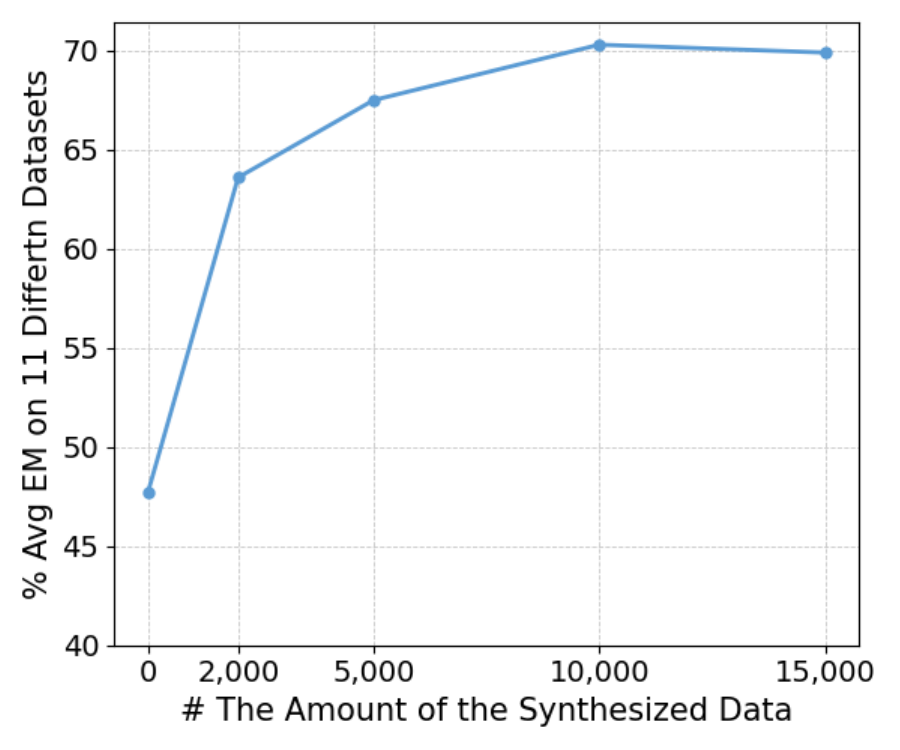}
    \caption{The Avg EM results (\%) on 11 datasets with different numbers of synthesized training data. We conduct the experiments based on LLaMA-3-Instruct-8B models.}
    \label{fig_amout}
\end{figure}

\noindent
\textbf{Multilingual Transfer Ability and Context Length Generalization of \textsc{Canoe}.}
To further explore the multilingual transfer ability of \textsc{Canoe}, we further evaluate our model on the Chinese dataset.
Specifically, we use the single-document QA dataset MultiFieldQA-zh \citep{bai2023longbench}, the multi-document QA dataset DuReader \citep{he-etal-2018-dureader}, and the summarization dataset VCSUM \citep{wu-etal-2023-vcsum} within LongBench \citep{bai2023longbench}.
Following \citet{si-etal-2025-gateau} that utilizes the GPT-4 to evaluate the correctness of QA tasks and the faithfulness of the summarization task, we use the advanced LLM to evaluate these datasets.
We use the same prompts for three tasks as \citet{si-etal-2025-gateau} to query GPT-4o.
The test-time prompts for these tasks can be found in Figure \ref{fig:prompt_chinese_sqa}, Figure \ref{fig:prompt_chinese_mqa}, and Figure \ref{fig:prompt_chinese_sum}.
As shown in Table \ref{tab_chinese}, we can find that our proposed \textsc{Canoe} also improves the faithfulness in Chinese datasets, indicating that our proposed method has a strong multilingual transfer ability.
Meanwhile, these results also indicate that our method achieves better faithfulness even when our training data is short, i.e., \textsc{Canoe} guarantees a consistently strong performance in long-context scenarios.


\noindent
\textbf{Effect of the Amount of the Synthesized Short-form Data.}
To further explore the effect of the amount of the synthesized short-form data, we conduct the corresponding experiments as shown in Figure \ref{fig_amout}.
We can find that increasing the number of training data improves the performance of our method, but when the number of data is greater than 10,000, the performance will be stable.
For the models with different sizes, the optimal amount of training data may be different, e.g., the larger models may need more training data to achieve optimal performance.
Fortunately, due to our training data construction strategy, we can simply scale and synthesize training data without human annotation.

\begin{table}[t]
    \scriptsize
    \centering
    \resizebox{\linewidth}{!}{
        \begin{tabular}{m{\hsize}}
        
        \specialrule{1pt}{1pt}{2pt}

        \sethlcolor{tabcolor4}\hl{\textbf{Long-form QA}} \\
        Inputs:\{\} \\
        Question: What is the type and definition of this dance? \\
        
        \specialrule{0pt}{0pt}{1pt}
        \cdashline{1-1}[0.8pt/2pt]
        \specialrule{0pt}{0pt}{2pt}
        
        \sethlcolor{tabcolor6}\hl{\textbf{GRPO-LLaMA-8B}}: Ballet. (\xmark) \\

        \specialrule{0pt}{0pt}{1pt}
        \cdashline{1-1}[0.8pt/2pt]
        \specialrule{0pt}{0pt}{2pt}
        \sethlcolor{tabcolor1}\hl{\textbf{\textsc{Canoe}-LLaMA-8B}}: Based on the passage, ballet is a type of dance that developed from lavish court spectacles in Italy and France. It is a combination of music, drama, poetry, song, costumes, and dance, and involves a combination of different elements, including music, costumes, and performance. The passage also mentions that members of the court nobility took part as performers, and that professional dancers began to take the place of court amateurs. This suggests that ballet is a type of dance that is characterized by its combination of different art forms and its performance by trained dancers. (\cmark) \\
        \specialrule{1pt}{0pt}{1pt}
        
        \sethlcolor{tabcolor5}\hl{\textbf{Simplification}} \\
        Inputs: The Pokémon series began with the release of Pokémon Red and Green for the Game Boy in Japan. When these games proved extremely popular, an enhanced blue version was released sometime after, and the blue version was reprogrammed as Pokémon Red and Blue for international release. \\
        
        \specialrule{0pt}{0pt}{1pt}
        \cdashline{1-1}[0.8pt/2pt]
        \specialrule{0pt}{0pt}{2pt}
        
        \sethlcolor{tabcolor6}\hl{\textbf{GRPO-LLaMA-8B}}: Pocket Monsters Red and Green. (\xmark) \\

        \specialrule{0pt}{0pt}{1pt}
        \cdashline{1-1}[0.8pt/2pt]
        \specialrule{0pt}{0pt}{2pt}
        \sethlcolor{tabcolor1}\hl{\textbf{\textsc{Canoe}-LLaMA-8B}}: The Pokémon series started with the release of Pokémon Red and Green in Japan, which were later enhanced and reprogrammed for international release. (\cmark) \\
        
        \specialrule{1pt}{0pt}{1pt}
        \sethlcolor{tabcolor3}\hl{\textbf{Summarization}} \\ 
        Inputs: \{\} \\
        
        \specialrule{0pt}{0pt}{1pt}
        \cdashline{1-1}[0.8pt/2pt]
        \specialrule{0pt}{0pt}{2pt}
        
        \sethlcolor{tabcolor6}\hl{\textbf{GRPO-LLaMA-8B}}: Comedy clubs. (\xmark) \\

        \specialrule{0pt}{0pt}{1pt}
        \cdashline{1-1}[0.8pt/2pt]
        \specialrule{0pt}{0pt}{2pt}
        \sethlcolor{tabcolor1}\hl{\textbf{\textsc{Canoe}-LLaMA-8B}}:
        Kareem Badr and his friends were able to turn around the struggling comedy club, the Hideout, by applying their skills in problem-solving and efficiency. They broke down the business into small chunks and figured out how to make each aspect better and more efficient. This approach helped them to improve the club's performance and eventually expand it. The passage also mentions that the growing popularity of stand-up comedy is fueled in part by younger audiences, who view humorous TV hosts as their primary source for news. Many comedy clubs now require talent, and there is a career and a living to be made by good comedians who are not superstars. The clubs also make money from selling drinks and dinners, as well as from cover charges and talent fees. In addition, digital media has given live comedy a boost, with one terrific joke being able to get a million hits online.
        (\cmark) \\
        \specialrule{1pt}{0.5pt}{0pt}   
        \end{tabular}}
\caption{Case study between the models trained via GRPO and our model for long-form generation tasks.
Models are implemented on LLaMA-3-Instruct-8B.}
    
\label{tab:grpo_case}
\end{table}

\noindent
\textbf{Case Study between GRPO and the proposed Dual-GRPO.}
\label{appendix:grpo_vis}
We find that directly applying GRPO instead of our proposed Dual-GRPO to synthesized short-form data leads to over-optimizing short-form generation and a false response generation pattern.
The used system prompt for applying GRPO can be found in Figure \ref{fig:grpo_prompt}.
Shown in Table \ref{tab:grpo_case}, we can find that the tuned model GRPO-LLaMA-8B tends to directly copy text spans from the given context as the final answer instead of following instructions in long-form generation tasks.
However, when we apply Dual-GPRO to our synthesized data, we find that trained models can generate fluent and complete sentences.
Thus, Dual-GRPO not only improves the faithfulness of LLMs in two types of response generation but also ensures the utility of models.


\begin{table*}
\centering 
\scriptsize
\resizebox{0.7\columnwidth}{!}{%
\begin{tabular}{l|l}  
\toprule  
Relation & Description \\
\midrule  
P6 & head of government \\
P17 & country \\
P26 & spouse \\
P27 & country of citizenship \\
P30 & continent \\
P35 & head of state \\
P36 & capital \\
P37 & official language \\
P38 & currency \\
P39 & position held \\
P50 & author \\
P54 & member of sports team \\
P57 & director \\
P86 & composer \\
P101 & field of work \\
P103 & native language \\
P108 & employer \\
P112 & founder \\
P127 & owned by \\
P136 & genre \\
P1376 & capital of \\
P140 & religion \\
P155 & follows \\
P159 & headquarters location \\
P166 & award received \\
P170 & creator \\
P172 & ethnic group \\
P175 & performer \\
P178 & developer \\
P264 & record label \\
P276 & location \\
P286 & head coach \\
P407 & language of work or name \\
P413 & position played \\
P463 & member of \\
P488 & chairperson \\
P495 & country of origin \\
P641 & sport \\
P800 & notable work \\
P937 & work location \\
P169 & chief executive officer \\
\bottomrule  
\end{tabular}%
}  
\caption{Manually selected relations that are used to construct training data. We utilize the same relations as \citet{bi2024contextdpoaligninglanguagemodels}.}  
\label{tab:relations}  
\end{table*} 

\begin{figure*}
    \centering
    \begin{tcolorbox}[title = {Prompt for question generation for the samples with straightforward context.}, size=title, colframe = white, colbacktitle = black!65!white]
    [Instructions] \\
    You are a sophisticated question generator. 
    Given a triple \{$(h,r,t)$\} collected from Wikidata, generate a question that asks about the final tail entity \{$t$\} using the head entity \{$h$\} and the relation \{$r$\}. \\
    
    Directly give me the generated question:\\
    \end{tcolorbox}
    \caption{Prompt for question generation for the samples with straightforward context.}
    \label{fig:prompt_s_q}
\end{figure*}

\begin{figure*}
    \centering
    \begin{tcolorbox}[title = {Prompt for context generation for the samples with straightforward context.}, size=title, colframe = white, colbacktitle = black!65!white]
    [Instructions] \\
    You are a sophisticated context generator. 
    Given a triple \{$(h,r,t)$\} collected from Wikidata, generate a brief description of the head entity \{$h$\}, approximately 150 words long. Ensure the tail entity \{$t$\} and relation \{$r$\} are accurately mentioned in the generated description. \\
    
    Directly give me the generated context:\\
    \end{tcolorbox}
    \caption{Prompt for context generation for the samples with straightforward context.}
    \label{fig:prompt_s_c}
\end{figure*}

\begin{figure*}
    \centering
    \begin{tcolorbox}[title = {Prompt for question generation for the samples with reasoning-required context.}, size=title, colframe = white, colbacktitle = black!65!white]
    [Instructions] \\
    You are a sophisticated question generator. 
    Given a chain of triples \{[...]\} collected from Wikidata, generate a question that asks about the final tail entity \{$t$\} using the head entity \{$h$\} and the relation \{$r$\}. 
    Do not include any bridge entities in the question; instead, phrase the question as if directly asking about the relationship from the head entity to the tail entity \\
    
    Directly give me the generated question:\\
    \end{tcolorbox}
    \caption{Prompt for question generation for the samples with reasoning-required context.}
    \label{fig:prompt_r_q}
\end{figure*}

\begin{figure*}
    \centering
    \begin{tcolorbox}[title = {Prompt for context generation for the samples with reasoning-required context.}, size=title, colframe = white, colbacktitle = black!65!white]
    [Instructions] \\
    You are a sophisticated context generator. 
    Given a chain of triples \{[...]\} collected from Wikidata, generate a brief description of the head entity \{$h$\}, approximately \{150*n\} words long. 
    Ensure the tail entity \{$t$\} and relation \{$r$\} are accurately mentioned in the generated description. \\
    
    Directly give me the generated context:\\
    \end{tcolorbox}
    \caption{Prompt for context generation for the samples with reasoning-required context.}
    \label{fig:prompt_r_c}
\end{figure*}

\newpage

\begin{figure*}
    \centering
    \begin{tcolorbox}[title = {System prompt for Dual-GRPO.}, size=title, colframe = white, colbacktitle = black!65!white]
A conversation between User and Assistant. The user gives an instruction that consists of two parts: a passage and the actual instruction, separated by two newline characters. \\

The passage is provided within \textless context\textgreater and \textless /context\textgreater tags. The Assistant needs to refer to the given passage and complete the instruction. \\

The Assistant solves the question by first thinking about the reasoning process internally, according to the given passage, and then providing the response. \\

The response must be structured and include the following three sections, clearly marked by the respective tags: \\

\textbf{- Reasoning Process:} Explain your thought process or logical steps to derive the answer. Enclose this within \textless think\textgreater and \textless /think\textgreater tags. \\
\textbf{- Long Answer}: Provide a long response that consists of syntactically and semantically complete sentences to answer the question. 
Enclose this within \textless long\_answer\textgreater and \textless /long\_answer\textgreater tags. \\
\textbf{- Short Answer}: Present a concise response that directly answers the question. 
Enclose this within \textless short\_answer\textgreater and \textless /short\_answer\textgreater tags. \\

Format your response exactly as follows: \\
\textless think\textgreater reasoning process here. \textless /think\textgreater \textless long\_answer\textgreater detailed answer here. \textless /long\_answer\textgreater \textless short\_answer\textgreater the concise answer here. \textless /short\_answer\textgreater. \\
    \end{tcolorbox}
    \caption{System prompt for Dual-GRPO.}
    \label{fig:sys_prompt}
\end{figure*}

\clearpage


\begin{figure*}
    \centering
    \begin{tcolorbox}[title = {Prompt used to calculate quality score for text summarization.}, size=title, colframe = white, colbacktitle = black!65!white]
    You are asked to evaluate the quality of the AI assistant’s generated summary as an impartial judge, and your evaluation should take into account factors including readability (whether the summary is clear and easy to understand) and coherence (whether the assistant’s summary is logical and orderly). \\

    Read the AI assistant’s summary and input passages, and give an overall integer rating in on a scale of 1 to 5, where 1 is the lowest and 5 is the highest based on the evaluation criteria, strictly in the following format:``[[rating]]'', e.g. ``[[5]]''.\\

    Input Passages: \{\} \\
    Assistant’s summary:\{\} \\
    Rating: \\
    \end{tcolorbox}
    \caption{Prompt used to calculate quality score for text summarization.}
    \label{fig:prompt_scoring_sum}
\end{figure*}


\begin{figure*}
    \centering
    \begin{tcolorbox}[title = {Prompt used to calculate quality score for text simplification.}, size=title, colframe = white, colbacktitle = black!65!white]
    You are asked to evaluate the quality of the AI assistant’s generated text simplification as an impartial judge, and your evaluation should take into account factors including readability (whether the simplification is clear and easy to understand) and coherence (whether the assistant’s simplification is logical and orderly).  \\
    
    Read the AI assistant’s simplified version and the original text, and give an overall integer rating on a scale of 1 to 5, where 1 is the lowest and 5 is the highest based on the evaluation criteria, strictly in the following format: ``[[rating]]'', e.g. ``[[5]]''. \\
    
    Original text:  \{\} \\  
    AI assistant’s simplification:  \{\} \\  
    Rating: \\
    \end{tcolorbox}
    \caption{Prompt used to calculate quality score for text simplification.}
    \label{fig:prompt_scoring_sim}
\end{figure*}

\begin{figure*}
    \centering
    \begin{tcolorbox}[title = {Prompt used to calculate quality score for long-form QA.}, size=title, colframe = white, colbacktitle = black!65!white]
    You are asked to evaluate the quality of the AI assistant’s generated long-form answer as an impartial judge, and your evaluation should take into account factors including readability (whether the answer is clear and easy to understand) and coherence (whether the answer is logical and well-organized). \\
    
    Read the AI assistant’s long-form answer and the original question, and give an overall integer rating on a scale of 1 to 5, where 1 is the lowest and 5 is the highest, based on the evaluation criteria, strictly in the following format: ``[[rating]]'', e.g., ``[[5]]''. \\
    
    Question: \{\} \\
    Assistant’s long-form answer: \{\} \\
    Rating: \\
    \end{tcolorbox}
    \caption{Prompt used to calculate quality score for long-form QA.}
    \label{fig:prompt_scoring_lfqa}
\end{figure*}

\begin{figure*}
    \centering
    \begin{tcolorbox}[title = {Test-time prompt used for short-form QA tasks.}, size=title, colframe = white, colbacktitle = black!65!white]
    Passages: \{\} \\
    
    Refer to the passages above and answer the following question with just a few words. \\
    
    Question: \{\} \\
    
    Answer: \\
    \end{tcolorbox}
    \caption{Test-time prompt used for short-form QA tasks.}
    \label{fig:prompt_sfqa}
\end{figure*}

\begin{figure*}
    \centering
    \begin{tcolorbox}[title = {Test-time prompt used for multiple-choice QA task.}, size=title, colframe = white, colbacktitle = black!65!white]
    Passages: \{\} \\
    
    Refer to the passages above and answer the following question with just a few words. \\
    
    Question: \{\} \\

    Please select the correct option according to the question, and output the option letter (e.g. A/B/C/D): \\
    
    Options: \{\} \\
    
    Answer: \\
    \end{tcolorbox}
    \caption{Test-time prompt used for multiple-choice QA task.}
    \label{fig:prompt_mcqa}
\end{figure*}

\begin{figure*}
    \centering
    \begin{tcolorbox}[title = {Test-time prompt used for text summarization.}, size=title, colframe = white, colbacktitle = black!65!white]
    Passage: \{\} \\
    
    Refer to the passage above and provide a summary as the response. \\
    
    Summary: \\
    \end{tcolorbox}
    \caption{Test-time prompt used for text summarization.}
    \label{fig:prompt_sum}
\end{figure*}

\begin{figure*}
    \centering
    \begin{tcolorbox}[title = {Test-time prompt used for text simplification.}, size=title, colframe = white, colbacktitle = black!65!white]
    Passage: \{\} \\
    
    Refer to the passage above and simplify it to improve its readability, ensuring its core meaning remains intact. Please provide only the simplified text as the response. \\
    
    Simplified text: \\
    \end{tcolorbox}
    \caption{Test-time prompt used for text simplification.}
    \label{fig:prompt_sim}
\end{figure*}

\begin{figure*}
    \centering
    \begin{tcolorbox}[title = {Test-time prompt used for long-form QA task.}, size=title, colframe = white, colbacktitle = black!65!white]
    Passage: \{\} \\
    
    Refer to the passages above and answer the following question.  \\
    
    Question: \{ \} \\ 
    \end{tcolorbox}
    \caption{Test-time prompt used for long-form QA task.}
    \label{fig:prompt_lfqa}
\end{figure*}

\begin{figure*}
    \centering
    \begin{tcolorbox}[title = {Test-time prompt used for FaithEval in closed-book QA settings.}, size=title, colframe = white, colbacktitle = black!65!white]
    
    Question: \{\} \\

    Please select the correct option according to the question, and output the option letter (e.g. A/B/C/D): \\
    
    Options: \{\} \\
    
    Answer: \\
    \end{tcolorbox}
    \caption{Test-time prompt used for FaithEval in closed-book QA settings.}
    \label{fig:prompt_faitheval_fact}
\end{figure*}


\begin{figure*}
    \centering
    \begin{tcolorbox}[title = {The principles of human evaluation for long-form responses generation.}, size=title, colframe = white, colbacktitle = black!65!white]
    \noindent   
    You are asked to evaluate the responses generated by different models.
    You should choose the preferred responses according to the following perspectives independently: \\

    1. \textbf{Readability}: Whether the response is clear and easy to understand? \\
    
    2. \textbf{Faithfulness}: Whether the response is faithful to the context and the information can be grounded in the provided context. \\
    
    3. \textbf{Helpfulness}: Whether the response provides useful information and follows the instructions from users? \\
    
    4. \textbf{Naturalness}: Whether the response sounds natural and fluent? \\

    Finally, please make a decision among the 3 opinions, including Win, Tie, and Loss. \\

    \end{tcolorbox}
    \caption{The principles of human evaluation for long-form responses generation.}
    \label{fig:human_evaluation_principles}
\end{figure*}


\begin{figure*}
    \centering
    \begin{tcolorbox}[title = {System prompt for GRPO in the ablation study.}, size=title, colframe = white, colbacktitle = black!65!white]
A conversation between User and Assistant. The user gives an instruction that consists of two parts: a passage and the actual instruction, separated by two newline characters. \\

The passage is provided within \textless context\textgreater and \textless /context\textgreater tags. The Assistant needs to refer to the given passage and complete the instruction. \\

The Assistant solves the question by first thinking about the reasoning process internally, according to the given passage, and then providing the response. \\

The response must be structured and include the following two sections, clearly marked by the respective tags: \\

\textbf{- Reasoning Process:} Explain your thought process or logical steps to derive the answer. Enclose this within \textless think\textgreater and \textless /think\textgreater tags. \\
\textbf{- Answer}: Present a concise response that directly answers the question. 
Enclose this within \textless answer\textgreater and \textless /answer\textgreater tags. \\

Format your response exactly as follows: \\
\textless think\textgreater reasoning process here. \textless /think\textgreater \textless answer\textgreater answer here. \textless /answer\textgreater. \\
    \end{tcolorbox}
    \caption{System prompt for GRPO in the ablation study.}
    \label{fig:grpo_prompt}
\end{figure*}


\begin{figure*}
    \centering
    \begin{tcolorbox}[title = {Test-time prompt used for MultiField-zh.}, size=title, colframe = white, colbacktitle = black!65!white]

    \begin{CJK}{UTF8}{gbsn}
    阅读以下文字并用中文简短回答：\{\} \\
    现在请基于上面的文章回答下面的问题，只告诉我答案，不要输出任何其他字词。 \\
    问题：\{\}\\
    回答：\\
    \end{CJK}
    \end{tcolorbox}
    \caption{Test-time prompt used for MultiField-zh.}
    \label{fig:prompt_chinese_sqa}
\end{figure*}

\begin{figure*}
    \centering
    \begin{tcolorbox}[title = {Test-time prompt used for DuReader.}, size=title, colframe = white, colbacktitle = black!65!white]

    \begin{CJK}{UTF8}{gbsn}
    请基于给定的文章回答下述问题。 \\
    文章：\{\} \\
    问题：\{\} \\
    回答：\\
    \end{CJK}
    \end{tcolorbox}
    \caption{Test-time prompt used for DuReader.}
    \label{fig:prompt_chinese_mqa}
\end{figure*}

\begin{figure*}
    \centering
    \begin{tcolorbox}[title = {Test-time prompt used for VCSUM.}, size=title, colframe = white, colbacktitle = black!65!white]

    \begin{CJK}{UTF8}{gbsn}
    下面有一段会议记录，请你阅读后，写一段总结，总结会议的内容。 \\
    会议记录：\{\} \\
    会议总结：\\
    \end{CJK}
    \end{tcolorbox}
    \caption{Test-time prompt used for VCSUM.}
    \label{fig:prompt_chinese_sum}
\end{figure*}

\end{document}